# Sketching Meets Random Projection in the Dual: A Provable Recovery Algorithm for Big and High-dimensional Data

Jialei Wang*, Jason D. Lee♯, Mehrdad Mahdavi†, Mladen Kolar‡, and Nathan Srebro†

*Department of Computer Science, University of Chicago
♯Marshall School of Business, University of Southern California
†Toyota Technological Institute at Chicago
‡Booth School of Business, University of Chicago

**Abstract**

Sketching techniques have become popular for scaling up machine learning algorithms by reducing the sample size or dimensionality of massive data sets, while still maintaining the statistical power of big data. In this paper, we study sketching from an optimization point of view: we first show that the iterative Hessian sketch is an optimization process with *preconditioning*, and develop *accelerated* iterative Hessian sketch via the searching the conjugate direction; we then establish primal-dual connections between the Hessian sketch and dual random projection, and apply the preconditioned conjugate gradient approach on the dual problem, which leads to the *accelerated* iterative dual random projection methods. Finally to tackle the challenges from both large sample size and high-dimensionality, we propose the *primal-dual sketch*, which iteratively sketches the primal and dual formulations. We show that using a logarithmic number of calls to solvers of small scaled problem, primal-dual sketch is able to recover the optimum of the original problem up to *arbitrary* precision. The proposed algorithms are validated via extensive experiments on synthetic and real data sets which complements our theoretical results.

## 1 Introduction

Machine Learning has gained great empirical success from the massive data sets collected from various domains. Among them a major challenge is to utilize existing computational resources to build predictive and inferential models from such huge data sets, while maintaining the statistical power of big data. One remedy for the big data challenge is to build distributed computer systems and design distributed learning algorithms to make big data learning possible, however, distributed systems may not always available, and the cost of running distributed system can be much higher than one can afford, which makes distributed learning not suitable for all scenarios. An alternative remedy is to use the state-of-the-art randomized optimization algorithms to accelerate the training process, for example, researchers have proposed optimization algorithms for regularized empirical risk minimization problem, with provable fast convergence and low computational cost per iteration (see (Johnson and Zhang, 2013; Shalev-Shwartz and Zhang, 2013; Defazio et al., 2014) for examples), however, the speed of these optimization methods still heavily depends on the condition number of problem at hand, which can be undesirable for many real world problems.

Sketching (Woodruff, 2014), which approximates the solution via constructing some sketched, usually of smaller scale problem from the original data, has become an emerging technique for big data analytics. With the sketching technique, we can find solutions which *approximately* solve various forms of original large-scale problem, such as least square regression, robust regression, low-rank approximation, singular value decomposition, just to name a few. For survey and recent advances about sketching, we refer the readers to (Halko et al., 2011; Mahoney, 2011; Lu et al., 2013; Alaoui and Mahoney, 2014; Woodruff, 2014; Raskutti and Mahoney, 2015; Yang et al., 2015a; Oymak et al., 2015; Oymak and Tropp, 2015; Drineas and Mahoney, 2016) and references therein. However, one major drawback of sketching is that typically it's not suitable for the case if we want *high accurate* solution: to obtain a solution with exponentially smaller approximation error, we often need to increase the sketching dimension also *exponentially*.

The situation has become better with recent work on "iterative sketch", e.g. iterative Hessian sketch (IHS) (Pilanci and Wainwright, 2016) and iterative dual random projection (IDRP) (Zhang et al., 2014). These methods are able to refine their approximate solution by iteratively solving some small scale sketched problem. Among these innovations, Hessian sketch (Pilanci and Wainwright, 2016) is designed by reducing the sample size of the original problem, while dual random projection (Zhang et al., 2014) is proposed by reducing the dimension. As a consequence, when the sample size and feature dimension are both large, IHS and IDRP still need to solve relatively large-scale subproblems as they can only sketch the problem from one perspective.

In this paper, we make the following improvement upon previous work: we first propose an accelerated version of IHS which requires the same computational cost to solve the IHS subproblem at each sketching iteration, while with provably fewer number of sketching iterations to reach certain accuracy; we then reveal the primal-dual connections between IHS (Pilanci and Wainwright, 2016) and IDRP (Zhang et al., 2014), which are independently proposed by two different groups of researchers. In particular, we show that these two methods are equivalent in the sense that dual random projection is performing Hessian sketch in the dual space. Finally, to alleviate the computational issues raised by big and high-dimensional learning problems, we propose a *primal-dual* sketching method that can simultaneously reduce the sample size and dimension of the sketched sub-problem, with provable convergence guarantees.

**Organization** The rest of this paper is organized as follows: in Section 2 we review the iterative Hessian sketch as an optimization process and propose a new algorithm with faster convergence rate. In Section 3 we show that the dual random projection is equivalent to Hessian sketch, and propose the corresponding accelerated dual random projection as well. In Section 4 we combine the sketching from both primal and dual perspectives, and propose iterative algorithms by reduc-ing both sample size and problem dimension. We provide several theoretical analysis in Section 5, though we defer few technical results to the appendices, and conduct extensive experiments in Section 6. Finally we summarize and discuss several future directions in Section 7.

**Notation** We use bold-faced letters such as $\mathbf{w}$ to denote vectors, and bold-faced capital letters such as $\mathbf{X}$ to denote matrices. Given a matrix $\mathbf{X} \in \mathbb{R}^{n \times p}$, we define the following matrix induced norm for any vector $\mathbf{w} \in \mathbb{R}^p$,

$$\|\mathbf{w}\|_{\mathbf{X}} = \sqrt{\frac{\mathbf{w}^\top \mathbf{X}^\top \mathbf{X} \mathbf{w}}{n}}.$$

We use $\mathcal{N}(\boldsymbol{\mu}, \boldsymbol{\Sigma})$ to denote the multivariate normal distribution with mean $\boldsymbol{\mu}$ and covariance $\boldsymbol{\Sigma}$. We use $\mathbf{I}_n$ and $\mathbf{I}_p$ to denote the identity matrices of size $n \times n$ and $p \times p$, and $\lambda_{\max}(\mathbf{H})$ and $\lambda_{\min}(\mathbf{H})$ to denote the maximum and minimum eigenvalue of $\mathbf{H}$, respectively. For two sequences $\{a_n\}_{n=1}^\infty$ and $\{a_n\}_{n=1}^\infty$, we denote $a_n \lesssim b_n$ if $a_n \leq C b_n$ always holds for $n$ large enough with some constant $C$, and denote $a_n \gtrsim b_n$ if $b_n \lesssim a_n$. We also use the notation $a_n = \mathcal{O}(b_n)$ if $a_n \lesssim b_n$, and use $\widetilde{\mathcal{O}}(\cdot)$ for $\mathcal{O}(\cdot)$ to hide logarithmic factors.

## 2 Iterative Hessian Sketch as Optimization with Preconditioning

In this section, we first review the the iterative Hessian sketch proposed in (Pilanci and Wainwright, 2016) as an iterative preconditioned optimization process, and then propose a faster iterative algorithm by constructing better sketched problem to solve.

For ease of discussion, consider the following $\ell_2$ regularized least-squares (a.k.a. ridge regression) problem:

$$\min_{\mathbf{w} \in \mathbb{R}^p} P(\mathbf{X}, \mathbf{y}; \mathbf{w}) = \min_{\mathbf{w} \in \mathbb{R}^p} \frac{1}{2n} \|\mathbf{y} - \mathbf{X}\mathbf{w}\|_2^2 + \frac{\lambda}{2} \|\mathbf{w}\|_2^2. \tag{2.1}$$

where $\mathbf{X} \in \mathbb{R}^{n \times p}$ is the data matrix, $\mathbf{y} \in \mathbb{R}^n$ is the response vector. Let $\mathbf{w}^*$ denote the optimum of problem (2.1).

In real applications both $n$ and $p$ can be very large, thus sketching has become a widely used technique for finding an approximate solution of problem (2.1) efficiently (Drineas et al., 2011; Mahoney, 2011; Woodruff, 2014). In particular, to avoid solving a problem of huge sample size, the traditional sketching techniques (e.g. (Sarlos, 2006; Pilanci and Wainwright, 2015b)) were proposed to reduce the sample size from $n$ to $m$, where $m \ll n$, by solving the following sketched problem:

$$\min_{\mathbf{w} \in \mathbb{R}^p} P(\mathbf{\Pi}^\top \mathbf{X}, \mathbf{\Pi}^\top \mathbf{y}; \mathbf{w}) = \min_{\mathbf{w} \in \mathbb{R}^p} \frac{1}{2n} \left\|\mathbf{\Pi}^\top \mathbf{y} - \mathbf{\Pi}^\top \mathbf{X} \mathbf{w}\right\|_2^2 + \frac{\lambda}{2} \|\mathbf{w}\|_2^2, \tag{2.2}$$

where $\mathbf{\Pi} \in \mathbb{R}^{n \times m}$ is a sketching matrix, and typical choice of $\mathbf{\Pi}$ can be random Gaussian matrix, matrix with Rademacher entries, Sub-sampled Randomized Hadamard Transform (Boutsidis and Gittens, 2013) and Sub-sampled Randomized Fourier Transform (Rokhlin and Tygert, 2008), see discussions in Section 2.1 of (Pilanci and Wainwright, 2016) for details.

Though the classical sketching has been successful in various problems with provable guarantees, as shown in (Pilanci and Wainwright, 2016) there exists a approximation limit for the classical sketching methods to be practically useful, that is, to obtain an approximate solution with high precision, the sketching dimension $m$ needs to grow exponentially, which is impractical if we want a high accuracy approximation to the original problem, as the main purpose of sketching is to speed up the algorithms via reducing the sample size.

The main idea of Hessian sketch (Pilanci and Wainwright, 2016) is based on the following equivalent formulation of (2.1):

$$\min_{\mathbf{w} \in \mathbb{R}^p} P(\mathbf{X}, \mathbf{y}; \mathbf{w}) = \min_{\mathbf{w} \in \mathbb{R}^p} \frac{1}{2n} \|\mathbf{y}\|_2^2 + \frac{1}{2n} \|\mathbf{X}\mathbf{w}\|_2^2 - \frac{1}{n} \langle \mathbf{y}, \mathbf{X}\mathbf{w} \rangle + \frac{\lambda}{2} \|\mathbf{w}\|_2^2, \qquad (2.3)$$

and (Pilanci and Wainwright, 2016) proposed to only sketch the quadratic part $\|\mathbf{X}\mathbf{w}\|_2^2$ with respect to $\mathbf{X}$, but not the linear part $\langle \mathbf{y}, \mathbf{X}\mathbf{w} \rangle$. So the Hessian sketch considers solving the following sketched problem:

$$\min_{\mathbf{w} \in \mathbb{R}^p} P_{\text{HS}}(\mathbf{X}, \mathbf{y}; \mathbf{\Pi}, \mathbf{w}) = \min_{\mathbf{w} \in \mathbb{R}^p} \frac{1}{2n} \|\mathbf{y}\|_2^2 + \frac{1}{2n} \|\mathbf{\Pi}^\top \mathbf{X}\mathbf{w}\|_2^2 - \frac{1}{n} \langle \mathbf{y}, \mathbf{X}\mathbf{w} \rangle + \frac{\lambda}{2} \|\mathbf{w}\|_2^2. \qquad (2.4)$$

It is not hard to see that (2.4) has the following closed form solution:

$$\widehat{\mathbf{w}}_{\text{HS}} = \left( \lambda \mathbf{I}_p + \frac{\mathbf{X}^\top \mathbf{\Pi} \mathbf{\Pi}^\top \mathbf{X}}{n} \right)^{-1} \frac{\mathbf{X}^\top \mathbf{y}}{n}. \qquad (2.5)$$

We see that different from classical sketching where both data matrix $\mathbf{X}$ and the response vector $\mathbf{y}$ are both sketched, in Hessian sketch the only sketched part is the Hessian matrix, through the following transform:

$$\mathbf{X}^\top \mathbf{X} \to \mathbf{X}^\top \mathbf{\Pi} \mathbf{\Pi}^\top \mathbf{X}.$$

Though the Hessian sketch suffers from the same approximation limit as the classical sketch, one notable feature of Hessian sketch is that one can implement an iterative extension to refine the approximation to higher precision. To this end, define the initial Hessian sketch approximation as $\widehat{\mathbf{w}}_{\text{HS}}^{(1)}$:

$$\widehat{\mathbf{w}}_{\text{HS}}^{(1)} = \arg\min_{\mathbf{w}} \mathbf{w}^\top \left( \frac{\mathbf{X}^\top \mathbf{\Pi} \mathbf{\Pi}^\top \mathbf{X}}{2n} + \frac{\lambda}{2} \mathbf{I}_p \right) \mathbf{w} - \frac{1}{n} \langle \mathbf{y}, \mathbf{X}\mathbf{w} \rangle.$$

After obtaining $\widehat{\mathbf{w}}_{\text{HS}}^{(1)}$, we can consider the following optimization problem:

$$\arg\min_{\mathbf{u}} \frac{1}{2n} \left\| \mathbf{y} - \mathbf{X}(\mathbf{u} + \widehat{\mathbf{w}}_{\text{HS}}^{(1)}) \right\|_2^2 + \frac{\lambda}{2} \left\| (\mathbf{u} + \widehat{\mathbf{w}}_{\text{HS}}^{(1)}) \right\|_2^2$$
$$= \arg\min_{\mathbf{u}} \mathbf{u}^\top \left( \frac{\mathbf{X}^\top \mathbf{X}}{2n} + \frac{\lambda}{2} \mathbf{I}_p \right) \mathbf{u} - \left\langle \frac{\mathbf{X}^\top (\mathbf{y} - \mathbf{X}\widehat{\mathbf{w}}_{\text{HS}}^{(t)})}{n} - \lambda \widehat{\mathbf{w}}_{\text{HS}}^{(t)}, \mathbf{u} \right\rangle.$$

It is clear that $\mathbf{w}^* - \widehat{\mathbf{w}}_{\text{HS}}^{(1)}$ is the optimum for above problem. The main idea of iterative Hessian sketch (IHS) is to approximate the residual $\mathbf{w}^* - \widehat{\mathbf{w}}_{\text{HS}}^{(1)}$ by Hessian sketch again. At time $t$, let $\mathbf{u}^{(t)}$ be the approximate of $\mathbf{w}^* - \widehat{\mathbf{w}}_{\text{HS}}^{(t)}$ via solving the following sketched problem:

$$\widehat{\mathbf{u}}^{(t)} = \arg\min_{\mathbf{u}} \mathbf{u}^\top \left( \frac{\mathbf{X}^\top \mathbf{\Pi} \mathbf{\Pi}^\top \mathbf{X}}{2n} + \frac{\lambda}{2} \mathbf{I}_p \right) \mathbf{u} - \left\langle \frac{\mathbf{X}^\top (\mathbf{y} - \mathbf{X}\widehat{\mathbf{w}}_{\text{HS}}^{(t)})}{n} - \lambda \widehat{\mathbf{w}}_{\text{HS}}^{(t)}, \mathbf{u} \right\rangle, \qquad (2.6)$$

---
**Algorithm 1:** Iterative Hessian Sketch (IHS).
---
1 **Input:** Data $\mathbf{X}, \mathbf{y}$, sketching matrix $\mathbf{\Pi}$.
2 **Initialization:** $\widehat{\mathbf{w}}_{\text{HS}}^{(0)} = \mathbf{0}$.
3 **for** $t = 0, 1, 2, \ldots$ **do**
4 $\quad$ Update the approximation by $\widehat{\mathbf{w}}_{\text{HS}}^{(t+1)} = \widehat{\mathbf{w}}_{\text{HS}}^{(t)} + \widehat{\mathbf{u}}^{(t)}$, where $\widehat{\mathbf{u}}^{(t)}$ is obtained by solving the sketched problem (2.6).
5 **end**
---

and we then update $\widehat{\mathbf{w}}^{(t+1)}$ by
$$\widehat{\mathbf{w}}_{\text{HS}}^{(t+1)} = \widehat{\mathbf{w}}_{\text{HS}}^{(t)} + \widehat{\mathbf{u}}^{(t)}.$$

The algorithm for IHS is shown in Algorithm 1. Since (2.6) is a sketched problem with sample size $m$, it can be solved more efficiently than the original problem (2.1). Besides, we can reuse the previous sketched data $\mathbf{\Pi}^\top \mathbf{X}$ without constructing any new random sketching matrix. Moreover, (Pilanci and Wainwright, 2016) showed that the approximation error of IHS is exponentially decreasing when we increase the number of sketching iterations, thus IHS could find an approximated solution with $\epsilon$-approximation error within $\mathcal{O}(\log(1/\epsilon))$ iterations, as long as the sketching dimension $m$ is large enough. Moreover, though this powerful technique is originally focused on the least-squares problem (2.1), the idea of IHS can be extended to solve more general problems, such as constrained least-squares (Pilanci and Wainwright, 2016), optimization with self-concordant loss (Pilanci and Wainwright, 2015a), as well as non-parametric methods (Yang et al., 2015b).

Though IHS improved the classical sketching by enabling us to find an high quality approximation more efficiently, it is imperfect due to the following reasons:

- The "exponentially approximation error decreasing" guarantee relies on the basis that the sketching dimension $m$ is large enough. The necessary sketching dimension depends on the intrinsic complexity of the problem, however, if the "sufficient sketching dimension" condition is violated, as we will show in experiments, IHS can even diverge, i.e. we obtain arbitrary worse approximation after applying IHS.

- As we will show later, even when the "sufficient sketching dimension" condition is satisfied, the decreasing speed of the approximation error in IHS can be significantly improved.

Here, we show that the iterative Hessian sketch is in fact an optimization process with *preconditioning*. For notation simplicity let

$$\mathbf{H} = \frac{\mathbf{X}^\top \mathbf{X}}{n} + \lambda \mathbf{I}_p \quad \text{and} \quad \widetilde{\mathbf{H}} = \frac{\mathbf{X}^\top \mathbf{\Pi} \mathbf{\Pi}^\top \mathbf{X}}{n} + \lambda \mathbf{I}_p,$$

and we use the following gradient notion on $P(\mathbf{X}, y; \mathbf{w})$:

$$\nabla P(\mathbf{w}) = -\frac{\mathbf{X}^\top (\mathbf{y} - \mathbf{X}\mathbf{w})}{n} + \lambda \mathbf{w}.$$

Then it is not hard to see that the IHS in Algorithm 1 is performing the following iterative update:

$$\widehat{\mathbf{w}}_{\text{HS}}^{(t+1)} = \widehat{\mathbf{w}}_{\text{HS}}^{(t)} - \widetilde{\mathbf{H}}^{-1}\nabla P(\widehat{\mathbf{w}}_{\text{HS}}^{(t)}),$$

which is like a Newton update where we replace the true Hessian $\mathbf{H}$ with the sketched Hessian $\widetilde{\mathbf{H}}$. This update can be derived by using the change of variable $\mathbf{z} = \widetilde{\mathbf{H}}^{1/2}\mathbf{w}$, and then applying gradient descent in the $\mathbf{z}$ space:

$$\widehat{\mathbf{z}}^{(t+1)} = \widehat{\mathbf{z}}^{(t)} - \nabla_{\mathbf{z}} P(\widetilde{\mathbf{H}}^{-1/2}\mathbf{z}) = \widehat{\mathbf{z}}^{(t)} - \widetilde{\mathbf{H}}^{-1/2}\nabla_{\mathbf{x}} P(\widetilde{\mathbf{H}}^{-1/2}\widehat{\mathbf{z}}^{(t)}).$$

By changing the update back to the original space by multiplying $\widetilde{\mathbf{H}}^{-1/2}$, we get back the IHS update:

$$\widehat{\mathbf{w}}_{\text{HS}}^{(t+1)} = \widehat{\mathbf{w}}_{\text{HS}}^{(t)} - \widetilde{\mathbf{H}}^{-1}\nabla P(\widehat{\mathbf{w}}_{\text{HS}}^{(t)}).$$

With above discussion, we see that the iterative Hessian sketch is in fact an optimization process with the sketched Hessian as preconditioning.

## 2.1 Accelerated IHS via Preconditioned Conjugate Gradient

In this section, we present the accelerated iterative Hessian sketch (Acc-IHS) algorithm by utilizing the idea of preconditioned conjugate gradient. Conjugate gradient is known to have better convergence properties than gradient descent in solving linear systems (Hestenes and Stiefel, 1952; Nocedal and Wright, 2006). Since iterative Hessian sketch is gradient descent in the transformed space $\mathbf{z} = \widetilde{\mathbf{H}}^{1/2}\mathbf{w}$, this suggests accelerating by performing conjugate gradient in the transformed space [1], instead of gradient descent.

This leads to the algorithm Acc-IHS as detailed in Algorithm 2, where at each iteration we call the solver for the following sketched linear system:

$$\widehat{\mathbf{u}}^{(t)} = \arg\min_{\mathbf{u}} \mathbf{u}^\top \left( \frac{\mathbf{X}^\top \mathbf{\Pi} \mathbf{\Pi}^\top \mathbf{X}}{2n} + \frac{\lambda}{2}\mathbf{I}_p \right) \mathbf{u} - \left\langle \mathbf{r}^{(t)}, \mathbf{u} \right\rangle. \tag{2.7}$$

Unlike IHS which uses $\widetilde{\mathbf{H}}^{-1}\nabla P(\widehat{\mathbf{w}}_{\text{HS}}^{(t)})$ as the update direction at time $t$, for Acc-IHS we use $\mathbf{p}^{(t)}$ as the update direction at time $t$, and $\mathbf{p}^{(t)}$ is chosen to satisfy the conjugate condition:

$$\left(\mathbf{p}^{(t_1)}\right)^\top \left( \frac{\mathbf{X}^\top \mathbf{\Pi} \mathbf{\Pi}^\top \mathbf{X}}{2n} + \frac{\lambda}{2}\mathbf{I}_p \right)^{-1} \left( \frac{\mathbf{X}^\top \mathbf{X}}{2n} + \frac{\lambda}{2}\mathbf{I}_p \right) \mathbf{p}^{(t_2)} = 0, \forall t_1, t_2 \geqslant 0, t_1 \neq t_2.$$

Since the updating direction is conjugate to the previous directions, it is guaranteed that after $p$ iterations, we reach the exact minimizer, i.e.,

$$\widehat{\mathbf{w}}_{\text{HS}}^{(t)} = \mathbf{w}^*, \forall t \geqslant p.$$

Moreover, Acc-IHS shares the same computational cost with standard IHS in solving each sketched subproblem, but as we will show in the theoretical analysis part, the convergence rate of Algorithm 2 is much faster than IHS, i.e., we need to solve much smaller number of sketched sub-problems than IHS to reach the same approximation accuracy.

---

[1]Equivalently, we can implicitly transform the space by defining inner product as $\langle \mathbf{x}, \mathbf{y} \rangle = \mathbf{x}^\top \widetilde{\mathbf{H}} \mathbf{y}$.

**Algorithm 2:** Accelerated Iterative Hessian Sketch (Acc-IHS).

1 **Input:** Data $\mathbf{X}, \mathbf{y}$, sketching matrix $\mathbf{\Pi}$.
2 **Initialization:** $\widehat{\mathbf{w}}_{\text{HS}}^{(0)} = \mathbf{0}, \mathbf{r}^{(0)} = -\frac{\mathbf{X}^\top \mathbf{y}}{n}$.
3 Compute $\widehat{\mathbf{u}}^{(0)}$ by solving (2.7), and update $\mathbf{p}^{(0)} = -\widehat{\mathbf{u}}^{(0)}$, calculate $\mathbf{v}^{(0)} = \left(\frac{\mathbf{X}^\top \mathbf{X}}{n} + \lambda \mathbf{I}_p\right) \mathbf{p}^{(0)}$.
4 **for** $t = 0, 1, 2, \ldots$ **do**
5     Calculate $\alpha^{(t)} = \frac{\langle \mathbf{r}^{(t)}, \mathbf{u}^{(t)} \rangle}{\langle \mathbf{p}^{(t)}, \mathbf{v}^{(t)} \rangle}$
6     Update the approximation by $\widehat{\mathbf{w}}_{\text{HS}}^{(t+1)} = \widehat{\mathbf{w}}_{\text{HS}}^{(t)} + \alpha^{(t)} \mathbf{p}^{(t)}$.
7     Update $\mathbf{r}^{(t+1)} = \mathbf{r}^{(t)} + \alpha^{(t)} \mathbf{v}^{(t)}$.
8     Update $\mathbf{u}^{(t+1)}$ by solving (2.7).
9     Update $\beta^{(t+1)} = \frac{\langle \mathbf{r}^{(t+1)}, \mathbf{u}^{(t)} \rangle}{\langle \mathbf{r}^{(t)}, \mathbf{r}^{(t)} \rangle}$.
10     Update $\mathbf{p}^{(t+1)} = -\mathbf{u}^{(t+1)} + \beta^{(t+1)} \mathbf{p}^{(t)}$.
11     Update $\mathbf{v}^{(t+1)} = \left(\frac{\mathbf{X}^\top \mathbf{X}}{n} + \lambda \mathbf{I}_p\right) \mathbf{p}^{(t+1)}$.
12 **end**

## 3 Equivalence between Dual Random Projection and Hessian Sketch

While Hessian sketch (Pilanci and Wainwright, 2016) tries to resolve the issue of huge sample size, Dual Random Projection (Zhang et al., 2013, 2014) is aimed to resolve the issue of high-dimensionality, where random projection as a standard technique is used to significantly reduce the dimension of data points. Again consider the standard ridge regression problem in (2.1), but now random projection is used to transform the original problem (2.1) to a low-dimensional problem:

$$\min_{\mathbf{w} \in \mathbb{R}^p} P_{\text{RP}}(\mathbf{XR}, \mathbf{y}; \mathbf{z}) = \min_{\mathbf{z} \in \mathbb{R}^d} \frac{1}{2n} \|\mathbf{y} - \mathbf{XRz}\|_2^2 + \frac{\lambda}{2} \|\mathbf{z}\|_2^2, \quad (3.1)$$

where $\mathbf{R} \in \mathbb{R}^{p \times d}$ is a random projection matrix, and $d \ll p$.

Let $\widehat{\mathbf{z}} = \arg\min_{\mathbf{z}} P_{\text{RP}}(\mathbf{XR}, \mathbf{y}; \mathbf{z})$. If we want to recover the original high-dimensional solution, (Zhang et al., 2014) observed that the naive recovered solution $\widehat{\mathbf{w}}_{\text{RP}} = \mathbf{R}\widehat{\mathbf{z}}$ is a bad approximation, and propose to recover $\mathbf{w}^*$ from the dual solution, which leads to the dual random projection (DRP) approach. To see this, consider the dual problem of the optimization problem in (2.1) as

$$\max_{\boldsymbol{\alpha} \in \mathbb{R}^n} D(\mathbf{X}, \mathbf{y}; \boldsymbol{\alpha}) = \max_{\boldsymbol{\alpha} \in \mathbb{R}^n} -\frac{1}{2n} \boldsymbol{\alpha}^\top \boldsymbol{\alpha} + \frac{\mathbf{y}^\top \boldsymbol{\alpha}}{n} - \frac{1}{2\lambda n^2} \boldsymbol{\alpha}^\top \mathbf{XX}^\top \boldsymbol{\alpha}. \quad (3.2)$$

Let $\boldsymbol{\alpha}^* = \arg\max_{\boldsymbol{\alpha} \in \mathbb{R}^n} D(\mathbf{X}, \mathbf{y}; \boldsymbol{\alpha})$ be the dual optimal solution. By standard primal-dual theory (Boyd and Vandenberghe, 2004), we have the following connection between the optimal primal and dual solutions:

$$\boldsymbol{\alpha}^* = \mathbf{y} - \mathbf{Xw}^* \quad \text{and} \quad \mathbf{w}^* = \frac{1}{\lambda n} \mathbf{X}^\top \boldsymbol{\alpha}^*. \quad (3.3)$$

The dual random projection procedure works as follows: first, we construct solve the low-dimensional, randomly projected problem (3.1) and obtain the solution $\widehat{\mathbf{z}}$, and then calculate the approximated

dual variables by:
$$\widehat{\boldsymbol{\alpha}}_{\text{DRP}} = \mathbf{y} - \mathbf{XR}\widehat{\mathbf{z}}, \tag{3.4}$$
based on the approximated dual solution $\widehat{\boldsymbol{\alpha}}_{\text{DRP}}$. Then we recover the primal solution as:
$$\widehat{\mathbf{w}}_{\text{DRP}} = \frac{1}{\lambda n}\mathbf{X}^\top \widehat{\boldsymbol{\alpha}}_{\text{DRP}}. \tag{3.5}$$
By combining above derivations, it is not hard to see that dual random projection for ridge regression has the following closed form solution:
$$\widehat{\mathbf{w}}_{\text{DRP}} = \frac{\mathbf{X}^\top}{n}\left(\lambda \mathbf{I}_n + \frac{\mathbf{XRR}^\top\mathbf{X}^\top}{n}\right)^{-1}\mathbf{y}. \tag{3.6}$$

In (Zhang et al., 2014) it has been shown that the recovered solution from dual, i.e. $\widehat{\mathbf{w}}_{\text{DRP}}$, is a much better approximation than the recovered solution directly from primal $\widehat{\mathbf{w}}_{\text{RP}}$. More specifically, they showed that $\widehat{\mathbf{w}}_{\text{RP}}$ is always a poor approximation of $\mathbf{w}^*$, because $\widehat{\mathbf{w}}_{\text{RP}}$ lives in a random subspace spanned by the random projection matrix $\mathbf{R}$. For $\widehat{\mathbf{w}}_{\text{DRP}}$, (Zhang et al., 2014) proved that as long as the projected dimension $d$ is large enough, $\widehat{\mathbf{w}}_{\text{DRP}}$ can be a good approximation of $\mathbf{w}^*$.

Moreover, (Zhang et al., 2014) proposed an iterative extension of DRP which can exponentially reducing the approximation error. To do so, suppose at iteration $t$ we have the approximate solution $\widehat{\mathbf{w}}_{\text{DRP}}^{(t)}$, and consider the following optimization problem:
$$\min_{\mathbf{u}\in\mathbb{R}^p} \frac{1}{2n}\left\|\mathbf{y} - \mathbf{X}(\mathbf{u} + \widehat{\mathbf{w}}_{\text{DRP}}^{(t)})\right\|_2 + \frac{\lambda}{2}\left\|\mathbf{u} + \widehat{\mathbf{w}}_{\text{DRP}}^{(t)}\right\|_2^2. \tag{3.7}$$
It is clear to see $\mathbf{w}^* - \widehat{\mathbf{w}}_{\text{DRP}}^{(t)}$ is the optimum solution of above optimization problem. The idea of iterative dual random projection (IDRP) is to approximate the residual $\mathbf{w}^* - \widehat{\mathbf{w}}_{\text{DRP}}^{(t)}$ by applying dual random projection again. That is, we construct the following randomly projected problem given $\widehat{\mathbf{w}}_{\text{DRP}}^{(t)}$:
$$\min_{\mathbf{z}\in\mathbb{R}^d} \frac{1}{2n}\left\|\mathbf{y} - \mathbf{Xw}_{\text{DRP}}^{(t)} - \mathbf{XRz}\right\|_2^2 + \frac{\lambda}{2}\left\|\mathbf{z} + \mathbf{R}^\top \mathbf{w}_{\text{DRP}}^{(t)}\right\|_2^2. \tag{3.8}$$
Let $\widehat{\mathbf{z}}^{(t)}$ to be the solution of (3.8), then we update the refined approximation of dual variables:
$$\widehat{\boldsymbol{\alpha}}_{\text{DRP}}^{(t+1)} = \mathbf{y} - \mathbf{Xw}_{\text{DRP}}^{(t)} - \mathbf{XR}\widehat{\mathbf{z}},$$
as well as the primal variables
$$\widehat{\mathbf{w}}_{\text{DRP}}^{(t+1)} = \frac{1}{\lambda n}\mathbf{X}^\top \widehat{\boldsymbol{\alpha}}_{\text{DRP}}^{(t+1)}.$$

The iterative dual random projection (IDRP) algorithm is shown in Algorithm 3. More generally, (Zhang et al., 2014) showed the iterative dual random projection can be used to solve any $\ell_2$ regularized empirical loss minimization problem as long as the loss function is smooth, typical examples include logistic regression, support vector machines with smoothed hinge loss, etc.

Though a powerful technique to cope with high-dimensionality issue, IDRP suffers from the same limitations as IHS: i) it requires the "large projection dimension" condition to make the approximation error decreasing, ii) the convergence speed of IDRP is not optimal. As will shown later, actually the dual random projection is equivalent to apply the Hessian sketch procedure on the dual problem, and we propose an accelerated IDRP approach to overcome above discussed limitations.

---
**Algorithm 3:** Iterative Dual Random Projection (IDRP).

1 **Input:** Data $\mathbf{X}, \mathbf{y}$, projection matrix $\mathbf{R}$.
2 **Initialization:** $\widehat{\mathbf{w}}_{\text{DRP}}^{(0)} = \mathbf{0}$.
3 **for** $t = 0, 1, 2, \ldots$ **do**
4  $\quad$ Solve the projected problem in (3.8) and obtain solution $\widehat{\mathbf{z}}^{(t)}$.
5  $\quad$ Update dual approximation: $\widehat{\boldsymbol{\alpha}}_{\text{DRP}}^{(t+1)} = \mathbf{y} - \mathbf{X}\mathbf{w}_{\text{DRP}}^{(t)} - \mathbf{X}\mathbf{R}\widehat{\mathbf{z}}^{(t)}$.
6  $\quad$ Update primal approximation: $\widehat{\mathbf{w}}_{\text{DRP}}^{(t+1)} = \frac{1}{\lambda n}\mathbf{X}^\top \widehat{\boldsymbol{\alpha}}_{\text{DRP}}^{(t+1)}$.
7 **end**
---

### 3.1 Dual Random Projection is Hessian Sketch in Dual Space

In this section we present one of the key observation, i.e., the equivalence between Hessian sketch and dual random projection. Note that the Hessian sketch is used for *sample reduction*, while the dual random projection is utilized for *dimension reduction*. Recall that the dual maximization objective (3.2) is a quadratic with respect to $\boldsymbol{\alpha}$, and we can write the equivalent minimization objective in the following form:

$$\min_{\boldsymbol{\alpha}\in\mathbb{R}^n} \boldsymbol{\alpha}^\top \left(\frac{\mathbf{X}\mathbf{X}^\top}{2\lambda n} + \frac{1}{2}\mathbf{I}_n\right)\boldsymbol{\alpha} - \langle \mathbf{y}, \boldsymbol{\alpha}\rangle. \tag{3.9}$$

We can treat (3.9) as our primal problem and applying the Hessian sketch with sketching matrix $\mathbf{R} \in \mathbb{R}^{p\times d}$ to find an approximated solution for $\boldsymbol{\alpha}^*$:

$$\widehat{\boldsymbol{\alpha}}_{\text{HS}} = \arg\min_{\boldsymbol{\alpha}\in\mathbb{R}^n} \boldsymbol{\alpha}^\top \left(\frac{\mathbf{X}\mathbf{R}\mathbf{R}^\top\mathbf{X}^\top}{2\lambda n} + \frac{1}{2}\mathbf{I}_n\right)\boldsymbol{\alpha} - \langle \mathbf{y}, \boldsymbol{\alpha}\rangle, \tag{3.10}$$

which has the closed form solution as

$$\widehat{\boldsymbol{\alpha}}_{\text{HS}} = \lambda \left(\lambda \mathbf{I}_n + \frac{\mathbf{X}\mathbf{R}\mathbf{R}^\top\mathbf{X}^\top}{n}\right)^{-1}\mathbf{y}.$$

If we substitute $\widehat{\boldsymbol{\alpha}}_{\text{HS}}$ to the primal-dual connection (3.3), we obtained an approximated primal solution:

$$\widehat{\mathbf{w}} = \frac{\mathbf{X}^\top}{n}\left(\lambda\mathbf{I}_n + \frac{\mathbf{X}\mathbf{R}\mathbf{R}^\top\mathbf{X}^\top}{n}\right)^{-1}\mathbf{y}.$$

Compared with the DRP approximation (3.6), we see these two approximations coincident, thus we see that *Dual Random Projection is Hessian sketch applied in dual space*. For ridge regression

**Algorithm 4:** Accelerated Iterative Dual Random Projection (Acc-IDRP)—Primal Version.

1 **Input:** Data $\mathbf{X}, \mathbf{y}$, projection matrix $\mathbf{R}$.
2 **Initialization:** $\widehat{\mathbf{w}}_{\text{DRP}}^{(0)} = \mathbf{0}, \widehat{\boldsymbol{\alpha}}_{\text{DRP}}^{(0)} = \mathbf{0}, \mathbf{r}^{(0)} = -\mathbf{y}$.
3 Compute $\mathbf{z}^{(0)}$ by solving (3.11), and update $\mathbf{u}^{(0)} = \mathbf{r}^{(0)} - \mathbf{X}\mathbf{R}\mathbf{z}^{(0)}, \mathbf{p}^{(0)} = -\mathbf{u}^{(0)}$,
$\mathbf{v}^{(0)} = \left(\frac{\mathbf{X}\mathbf{X}^\top}{n} + \lambda \mathbf{I}_n\right)\mathbf{p}^{(0)}$.
4 **for** $t = 0, 1, 2, \ldots$ **do**
5     Calculate $a^{(t)} = \frac{\langle \mathbf{r}^{(t)}, \mathbf{u}^{(t)} \rangle}{\langle \mathbf{p}^{(t)}, \mathbf{v}^{(t)} \rangle}$
6     Update the dual approximation by $\widehat{\boldsymbol{\alpha}}_{\text{DRP}}^{(t+1)} = \widehat{\boldsymbol{\alpha}}_{\text{DRP}}^{(t)} + a^{(t)}\mathbf{p}^{(t)}$.
7     Update primal approximation: $\widehat{\mathbf{w}}_{\text{DRP}}^{(t+1)} = \frac{1}{\lambda n}\mathbf{X}^\top \widehat{\boldsymbol{\alpha}}_{\text{DRP}}^{(t+1)}$.
8     Update $\mathbf{r}^{(t+1)} = \mathbf{r}^{(t)} + a^{(t)}\mathbf{v}^{(t)}$.
9     Solve the projected problem in (3.11) and obtain solution $\widehat{\mathbf{z}}^{(t+1)}$.
10     Update $\mathbf{u}^{(t+1)} = \mathbf{r}^{(t+1)} - \mathbf{X}\mathbf{R}\widehat{\mathbf{z}}^{(t+1)}$.
11     Update $\beta^{(t+1)} = \frac{\langle \mathbf{r}^{(t+1)}, \mathbf{u}^{(t)} \rangle}{\langle \mathbf{r}^{(t)}, \mathbf{r}^{(t)} \rangle}$.
12     Update $\mathbf{p}^{(t+1)} = -\mathbf{u}^{(t+1)} + \beta^{(t+1)}\mathbf{p}^{(t)}$.
13     Update $\mathbf{v}^{(t+1)} = \left(\frac{\mathbf{X}\mathbf{X}^\top}{n} + \lambda \mathbf{I}_n\right)\mathbf{p}^{(t+1)}$.
14 **end**

problem (2.1) one have closed form solutions for various sketching techniques as:

$$\textbf{Original}: \quad \mathbf{w}^* = \left(\lambda \mathbf{I}_p + \frac{\mathbf{X}^\top \mathbf{X}}{n}\right)^{-1} \frac{\mathbf{X}^\top \mathbf{y}}{n} = \frac{\mathbf{X}^\top}{n}\left(\lambda \mathbf{I}_n + \frac{\mathbf{X}\mathbf{X}^\top}{n}\right)^{-1}\mathbf{y}$$

$$\textbf{Classical Sketch}: \quad \widehat{\mathbf{w}}_{\text{CS}} = \left(\lambda \mathbf{I}_p + \frac{\mathbf{X}^\top \mathbf{\Pi}\mathbf{\Pi}^\top \mathbf{X}}{n}\right)^{-1} \frac{\mathbf{X}^\top \mathbf{\Pi}\mathbf{\Pi}^\top \mathbf{y}}{n}$$

$$\textbf{Random Projection}: \quad \widehat{\mathbf{w}}_{\text{RP}} = \mathbf{R}\left(\lambda \mathbf{I}_d + \frac{\mathbf{R}^\top \mathbf{X}^\top \mathbf{X}\mathbf{R}}{n}\right)^{-1}\mathbf{R}^\top \frac{\mathbf{X}^\top \mathbf{y}}{n}$$

$$\textbf{Hessian Sketch}: \quad \widehat{\mathbf{w}}_{\text{HS}} = \left(\lambda \mathbf{I}_p + \frac{\mathbf{X}^\top \mathbf{\Pi}\mathbf{\Pi}^\top \mathbf{X}}{n}\right)^{-1} \frac{\mathbf{X}^\top \mathbf{y}}{n}$$

$$\textbf{Dual Random Projection}: \quad \widehat{\mathbf{w}}_{\text{DRP}} = \frac{\mathbf{X}^\top}{n}\left(\lambda \mathbf{I}_n + \frac{\mathbf{X}\mathbf{R}\mathbf{R}^\top \mathbf{X}^\top}{n}\right)^{-1}\mathbf{y}$$

As we can see above, Hessian sketch is essentially sketching the *covariance matrix*:

$$\mathbf{X}^\top \mathbf{X} \to \mathbf{X}^\top \mathbf{\Pi}\mathbf{\Pi}^\top \mathbf{X},$$

while DRP is essentially sketching the *Gram matrix*:

$$\mathbf{X}\mathbf{X}^\top \to \mathbf{X}\mathbf{R}\mathbf{R}^\top \mathbf{X}^\top.$$

### 3.2 Accelerated Iterative Dual Random Projection

Based on the equivalence between dual random projection and Hessian sketch established in Section 3.1, we proposed an accelerated iterative dual random projection algorithms which improves the convergence speed of standard iterative DRP procedure (Zhang et al., 2014). The algorithm is shown in Algorithm 4, in which at each iteration $t$, we call the solver for the following randomly projected problem based on the residual $\mathbf{r}^{(t)}$:

$$\widehat{\mathbf{z}}^{(t)} = \arg\min_{\mathbf{z}\in\mathbb{R}^d} \mathbf{z}^\top \left( \frac{\mathbf{R}^\top \mathbf{X}^\top \mathbf{X} \mathbf{R}}{2n} + \frac{\lambda}{2}\mathbf{I}_d \right) \mathbf{z} - \langle \mathbf{R}^\top \mathbf{X}^\top \mathbf{r}^{(t)}, \mathbf{z} \rangle. \quad (3.11)$$

The accelerated IDRP algorithms simulate the accelerated IHS algorithm (2), but run Acc-IHS in the dual space. However, Acc-IDRP is still an primal algorithm since it updates the corresponding dual variables back from solving the randomly projected primal problem (3.11). Algorithm 5 summarized the accelerated IDRP algorithms directly from the dual problem. We note that it is not a practical algorithm as it requires solving the relatively more expensive dual problem, however it is easier to understand as it directly borrows the ideas of Acc-IHS described in Section 25. For the dual version of Acc-IDRP algorithm, at each iteration it is required to solve the following dual optimization problem given the dual residual $\mathbf{r}^{(t)}$:

$$\widehat{\mathbf{u}}^{(t)} = \arg\min_{\mathbf{u}\in\mathbb{R}^n} \mathbf{u}^\top \left( \frac{\mathbf{X}\mathbf{R}\mathbf{R}^\top \mathbf{X}^\top}{2n} + \frac{\lambda}{2}\mathbf{I}_n \right) \mathbf{u} - \langle \mathbf{r}^{(t)}, \mathbf{u} \rangle. \quad (3.12)$$

As we will show later, though the computational cost per iteration of Acc-IDRP and standard IDRP is the same, Acc-IDRP has the following advantages over IDRP:

- As a preconditioned conjugate gradient procedure, Acc-IDRP is guaranteed to converge, and reach the optimum $\mathbf{w}^*$ within $n$ iterations, even when the projection dimension $d$ is very small.

- When the projection dimension $d$ is large enough to make standard IDRP converge quickly to the optimum, Acc-IDRP converges even faster.

## 4 Primal-Dual Sketch for Big and High-dimensional Problems

In this section, we combines the idea of iterative Hessian sketch and iterative dual random projection from the primal-dual point of view, and propose a more efficient sketching technique named *Iterative Primal-Dual Sketch* (IPDS) which simultaneously reduce the sample size and dimensionality of the problem, while still recovering the original solution to a high precision. Hessian sketch is particular suitable for the case where sample size is much larger than problem dimension, where the computational bottleneck is big $n$, and it is possible to use Hessian sketch to reduce the sample size significantly thus speed up the computation, and uses the iterative extension to reducing approximation error further to recover the original solution to a high precision; while dual random projection goes to another direction: it is mostly suitable for the case of high-dimensional data but relatively small sample size, where the computational bottleneck is "large $p$", and we could like to use random projection to perform dimension reduction thus gain speedup. Table 1 summarized these characteristics.

As shown in Table 1, Hessian sketch and dual random projection are suitable for case when the problem scales are not balanced, i.e. for the sample size and dimensions, one is much larger than

**Algorithm 5:** Accelerated Iterative Dual Random Projection (Acc-IDRP)—Dual Version.

1 **Input:** Data $\mathbf{X}, \mathbf{y}$, projection matrix $\mathbf{R}$.
2 **Initialization:** $\widehat{\mathbf{w}}_{\mathrm{DRP}}^{(0)} = \mathbf{0}, \widehat{\boldsymbol{\alpha}}_{\mathrm{DRP}}^{(0)} = \mathbf{0}, \mathbf{r}^{(0)} = -\mathbf{y}$.
3 Compute $\mathbf{u}^{(0)}$ by solving (3.12), and update $\mathbf{p}^{(0)} = -\mathbf{u}^{(0)}, \mathbf{v}^{(0)} = \left(\frac{\mathbf{X}\mathbf{X}^\top}{n} + \lambda \mathbf{I}_n\right)\mathbf{p}^{(0)}$.
4 **for** $t = 0, 1, 2, \ldots$ **do**
5 $\quad$ Calculate $a^{(t)} = \frac{\langle \mathbf{r}^{(t)}, \mathbf{u}^{(t)} \rangle}{\langle \mathbf{p}^{(t)}, \mathbf{v}^{(t)} \rangle}$
6 $\quad$ Update the dual approximation by $\widehat{\boldsymbol{\alpha}}_{\mathrm{DRP}}^{(t+1)} = \widehat{\boldsymbol{\alpha}}_{\mathrm{DRP}}^{(t)} + a^{(t)} \mathbf{p}^{(t)}$.
7 $\quad$ Update primal approximation: $\widehat{\mathbf{w}}_{\mathrm{DRP}}^{(t+1)} = \frac{1}{\lambda n} \mathbf{X}^\top \widehat{\boldsymbol{\alpha}}_{\mathrm{DRP}}^{(t+1)}$.
8 $\quad$ Update $\mathbf{r}^{(t+1)} = \mathbf{r}^{(t)} + a^{(t)} \mathbf{v}^{(t)}$.
9 $\quad$ Update $\mathbf{u}^{(t+1)}$ by solving (3.12).
10 $\quad$ Update $\beta^{(t+1)} = \frac{\langle \mathbf{r}^{(t+1)}, \mathbf{u}^{(t)} \rangle}{\langle \mathbf{r}^{(t)}, \mathbf{r}^{(t)} \rangle}$.
11 $\quad$ Update $\mathbf{p}^{(t+1)} = -\mathbf{u}^{(t+1)} + \beta^{(t+1)} \mathbf{p}^{(t)}$.
12 $\quad$ Update $\mathbf{v}^{(t+1)} = \left(\frac{\mathbf{X}\mathbf{X}^\top}{n} + \lambda \mathbf{I}_n\right)\mathbf{p}^{(t+1)}$.
13 **end**

| Approach | Suitable Situation | Reduced Dimension | Recovery | Iterative |
|---|---|---|---|---|
| `Classical Sketch` | large $n$, small $p$ | sample reduction | ✗ | ✗ |
| `Random Projection` | small $n$, large $p$ | dimension reduction | ✗ | ✗ |
| `Hessian Sketch` | large $n$, small $p$ | sample reduction | ✓ | ✓ |
| `DRP` | small $n$, large $p$ | dimension reduction | ✓ | ✓ |

Table 1: Comparison of various algorithms for data sketching in solving large-scale problems.

the other. For modern massive datasets, it is usually the case where both $n$ and $p$ are very large, for example, the click-through rate (CTR) prediction data sets provided by Criteo [2] has $n \geqslant 4 \times 10^9$ and $p \geqslant 8 \times 10^8$. Thus it is desirable to have a sketching method to simultaneously reduce the huge sample size and dimensionality.

Inspired by the primal-dual view described in Section 3.1, we propose the iterative Primal-Dual Sketch, which only involves solving small scale problems. For the original problem (2.1) with data $\{\mathbf{X}, \mathbf{y}\}$, we first construct the randomly projected data, as well as the *doubly sketched* data, as follows:

$$\mathbf{X} \to \mathbf{X}\mathbf{R}, \qquad \mathbf{X}\mathbf{R} \to \mathbf{\Pi}^\top \mathbf{X}\mathbf{R},$$

where $\mathbf{X}\mathbf{R}$ is the randomly projected data, and $\mathbf{\Pi}^\top \mathbf{X}\mathbf{R}$ is doubly sketched data by sketching $\mathbf{X}\mathbf{R}$ via sample reduction. We initialize the Primal-Dual Sketch solution as $\widehat{\mathbf{w}}_{\mathrm{DS}}^{(0)} = (\mathbf{0})$, and at every iteration, we first apply random projection on the primal problem (which is equivalent to Hessian Sketch on the dual problem), and obtain the following problem:

$$\min_{\mathbf{z} \in \mathbb{R}^d} \frac{1}{2n} \left\| \mathbf{y} - \mathbf{X}\widehat{\mathbf{w}}_{\mathrm{DS}}^{(t)} - \mathbf{X}\mathbf{R}\mathbf{z} \right\|_2^2 + \frac{\lambda}{2} \left\| \mathbf{z} + \mathbf{R}^\top \widehat{\mathbf{w}}_{\mathrm{DS}}^{(t)} \right\|_2^2, \qquad (4.1)$$

---
[2]http://labs.criteo.com/downloads/download-terabyte-click-logs/

**Algorithm 6:** Iterative Primal-Dual Sketch (IPDS).

**1 Input:** Data $\mathbf{X} \in \mathbb{R}^{n \times p}, \mathbf{y} \in \mathbb{R}^n$, sketching matrix $\mathbf{R} \in \mathbb{R}^{p \times d}, \mathbf{\Pi} \in \mathbb{R}^{n \times m}$.
**2 Initialization:** $\widehat{\mathbf{w}}_{\text{DS}}^{(0)} = \mathbf{0}$.
**3 for** $t = 0, 1, 2, \ldots$ **do**
**4**      **Initialization:** $\widetilde{\mathbf{z}}^{(0)} = \mathbf{0}, k = 0$
**5**      **if** *Not converged* **then**
**6**          Solve the sketched problem in (4.2) and obtain solution $\Delta \mathbf{z}^{(k)}$.
**7**          Update $\widetilde{\mathbf{z}}^{(k+1)} = \widetilde{\mathbf{z}}^{(k)} + \Delta \mathbf{z}^{(k)}$.
**8**          Update $k = k + 1$.
**9**      **end**
**10**      Update dual approximation: $\widehat{\boldsymbol{\alpha}}_{\text{DS}}^{(t+1)} = \mathbf{y} - \mathbf{X}\widehat{\mathbf{w}}_{\text{DS}}^{(t)} - \mathbf{X}\mathbf{R}\widetilde{\mathbf{z}}^{(k+1)}$.
**11**      Update primal approximation: $\widehat{\mathbf{w}}_{\text{DS}}^{(t+1)} = \frac{1}{\lambda n} \mathbf{X}^\top \widehat{\boldsymbol{\alpha}}_{\text{DS}}^{(t+1)}$.
**12 end**

which is the same as the iterative dual random projection subproblem (3.8). However, different fro IDRP, we don't directly solve (4.1), but apply the iterative Hessian sketch to find an approximate solution. Note that the expanded form of (4.1) is

$$\min_{\mathbf{z} \in \mathbb{R}^d} \mathbf{z}^\top \left( \frac{\mathbf{R}^\top \mathbf{X}^\top \mathbf{X} \mathbf{R}}{2n} + \frac{\lambda}{2} \mathbf{I}_d \right) \mathbf{z} - \left\langle \frac{(\mathbf{y} - \mathbf{X}\widehat{\mathbf{w}}_{\text{DS}}^{(t)})^\top \mathbf{X} \mathbf{R}}{n} - \lambda \mathbf{R}^\top \widehat{\mathbf{w}}_{\text{DS}}^{(t)}, \mathbf{z} \right\rangle.$$

We apply Hessian sketch to above problem, and iteratively refine the solution. That is, we first initialize an approximate solution of (4.1) of $\widetilde{\mathbf{z}}^{(0)}$ as $\mathbf{0}$, then at inner loop iteration $k$ find a solution for the following sketched problem:

$$\Delta \mathbf{z}^{(k)} = \arg\min_{\Delta \mathbf{z}} \Delta \mathbf{z}^\top \left( \frac{\mathbf{R}^\top \mathbf{X}^\top \mathbf{\Pi} \mathbf{\Pi}^\top \mathbf{X} \mathbf{R}}{2n} + \frac{\lambda}{2} \mathbf{I}_d \right) \mathbf{z}$$
$$- \left\langle \frac{\mathbf{R}^\top \mathbf{X}^\top (\mathbf{y} - \mathbf{X}\widehat{\mathbf{w}}_{\text{DS}}^{(t)} - \mathbf{X}\mathbf{R}\widetilde{\mathbf{z}}^{(k)})}{n} - \lambda \mathbf{R}^\top \widehat{\mathbf{w}}_{\text{DS}}^{(t)} - \lambda \widetilde{\mathbf{z}}^{(k)}, \Delta \mathbf{z} \right\rangle. \quad (4.2)$$

then update $\widetilde{\mathbf{z}}^{(k+1)}$ as

$$\widetilde{\mathbf{z}}^{(k+1)} = \widetilde{\mathbf{z}}^{(k)} + \Delta \mathbf{z}^{(k)}.$$

The key point is that for the subproblem (4.2), the sketched data matrix is only of size $m \times d$, compared to the original problem size $n \times p$, where $n \gg m, p \gg d$, in contrast, the IHS still need to solve sub-problems of size $m \times p$, while IDRP need to solve sub-problems of size $n \times d$. As will show in the theoretical analysis, we only need to call solvers of $m \times d$ problem (4.2) logarithmic times to obtain a solution of high approximation quality.

The pseudo code of Iterative Primal-Dual Sketch (IPDS) is summarized in Algorithm 6. It is also possible to perform iterative Primal-Dual Sketch via another direction, that is, first perform primal Hessian sketch, and then apply dual Hessian sketch to solve the sketched primal problem:

$$\mathbf{X} \to \mathbf{\Pi}^\top \mathbf{X}, \quad \mathbf{\Pi}^\top \mathbf{X} \to \mathbf{\Pi}^\top \mathbf{X} \mathbf{R}.$$

**Algorithm 7:** Accelerated Iterative Primal-Dual Sketch (Acc-IPDS).

1 **Input:** Data $\mathbf{X} \in \mathbb{R}^{n \times p}, \mathbf{y} \in \mathbb{R}^n$, sketching matrix $\mathbf{R} \in \mathbb{R}^{p \times d}, \mathbf{\Pi} \in \mathbb{R}^{n \times m}$.
2 **Initialization:** $\widehat{\mathbf{w}}_{\text{DS}}^{(0)} = \mathbf{0}, \widehat{\boldsymbol{\alpha}}_{\text{DS}}^{(0)} = \mathbf{0}, \mathbf{r}_{\text{Dual}}^{(0)} = -\mathbf{y}$.
3 **Initialization:** $k = 0, \widetilde{\mathbf{z}}^{(k)} = \mathbf{0}, \mathbf{r}_{\text{P}}^{(0)} = \mathbf{R}^\top \mathbf{X}^\top \mathbf{r}_{\text{D}}^{(0)}$.
4 Compute $\widehat{\mathbf{u}}_{\text{P}}^{(0)}$ by solving (4.3), and update $\mathbf{p}_{\text{P}}^{(0)} = -\mathbf{u}_{\text{P}}^{(0)}$, calculate
$\mathbf{v}_{\text{P}}^{(0)} = \left( \frac{\mathbf{R}^\top \mathbf{X} \mathbf{X} \mathbf{R}}{n} + \lambda \mathbf{I}_d \right) \mathbf{p}_{\text{P}}^{(0)}$.
5 **if** *Not converged* **then**
6 $\quad$ Calculate $a_{\text{P}}^{(k)} = \frac{\langle \mathbf{r}_{\text{P}}^{(k)}, \mathbf{u}_{\text{P}}^{(k)} \rangle}{\langle \mathbf{p}_{\text{P}}^{(k)}, \mathbf{v}_{\text{P}}^{(k)} \rangle}$, and update the approximation by $\widetilde{\mathbf{z}}^{(k+1)} = \widetilde{\mathbf{z}}^{(k)} + a_{\text{P}}^{(k)} \mathbf{p}_{\text{P}}^{(k)}$.
7 $\quad$ Update $\mathbf{r}_{\text{P}}^{(k+1)} = \mathbf{r}_{\text{P}}^{(k)} + a_{\text{P}}^{(k)} \mathbf{v}^{(k)}$, and update $\mathbf{u}_{\text{P}}^{(k+1)}$ by solving (4.3).
8 $\quad$ Update $\beta_{\text{P}}^{(k+1)} = \frac{\langle \mathbf{r}_{\text{P}}^{(k+1)}, \mathbf{u}_{\text{P}}^{(k)} \rangle}{\langle \mathbf{r}_{\text{P}}^{(k)}, \mathbf{r}_{\text{P}}^{(k)} \rangle}$, and update $\mathbf{p}_{\text{P}}^{(k+1)} = -\mathbf{u}_{\text{P}}^{(k+1)} + \beta_{\text{P}}^{(k+1)} \mathbf{p}_{\text{P}}^{(k)}$.
9 $\quad$ Update $\mathbf{v}_{\text{P}}^{(k+1)} = \left( \frac{\mathbf{R}^\top \mathbf{X}^\top \mathbf{X} \mathbf{R}}{n} + \lambda \mathbf{I}_p \right) \mathbf{p}_{\text{P}}^{(t+1)}$, and update $k = k + 1$.
10 **end**
11 Compute $\mathbf{u}_{\text{D}}^{(0)} = \mathbf{r}_{\text{D}}^{(0)} - \mathbf{X} \mathbf{R} \widetilde{\mathbf{z}}^{(k+1)}, \mathbf{p}_{\text{D}}^{(0)} = -\mathbf{u}_{\text{D}}^{(0)}, \mathbf{v}_{\text{D}}^{(0)} = \left( \frac{\mathbf{X} \mathbf{X}^\top}{n} + \lambda \mathbf{I}_n \right) \mathbf{p}_{\text{D}}^{(0)}$.
12 **for** $t = 0, 1, 2, \ldots$ **do**
13 $\quad$ Calculate $a_{\text{D}}^{(t)} = \frac{\langle \mathbf{r}_{\text{D}}^{(t)}, \mathbf{u}_{\text{D}}^{(t)} \rangle}{\langle \mathbf{p}_{\text{D}}^{(t)}, \mathbf{v}_{\text{D}}^{(t)} \rangle}$, and update the dual approximation by $\widehat{\boldsymbol{\alpha}}_{\text{DS}}^{(t+1)} = \widehat{\boldsymbol{\alpha}}_{\text{DS}}^{(t)} + a_{\text{D}}^{(t)} \mathbf{p}_{\text{D}}^{(t)}$.
14 $\quad$ Update primal approximation: $\widehat{\mathbf{w}}_{\text{DS}}^{(t+1)} = \frac{1}{\lambda n} \mathbf{X}^\top \widehat{\boldsymbol{\alpha}}_{\text{DS}}^{(t+1)}$, and update $\mathbf{r}_{\text{D}}^{(t+1)} = \mathbf{r}_{\text{D}}^{(t)} + a_{\text{D}}^{(t)} \mathbf{v}_{\text{D}}^{(t)}$.
15 $\quad$ **Initialization:** $k = 0, \widetilde{\mathbf{z}}^{(k)} = \mathbf{0}, \mathbf{r}_{\text{P}}^{(0)} = \mathbf{R}^\top \mathbf{X}^\top \mathbf{r}_{\text{D}}^{(t+1)}$.
16 $\quad$ Compute $\widehat{\mathbf{u}}_{\text{P}}^{(0)}$ by solving (4.3), and update $\mathbf{p}_{\text{P}}^{(0)} = -\mathbf{u}_{\text{P}}^{(0)}$, calculate
$\mathbf{v}_{\text{P}}^{(0)} = \left( \frac{\mathbf{R}^\top \mathbf{X} \mathbf{X} \mathbf{R}}{n} + \lambda \mathbf{I}_d \right) \mathbf{p}_{\text{P}}^{(0)}$.
17 $\quad$ **if** *Not converged* **then**
18 $\quad\quad$ Calculate $a_{\text{P}}^{(k)} = \frac{\langle \mathbf{r}_{\text{P}}^{(k)}, \mathbf{u}_{\text{P}}^{(k)} \rangle}{\langle \mathbf{p}_{\text{P}}^{(k)}, \mathbf{v}_{\text{P}}^{(k)} \rangle}$, and update the approximation by $\widetilde{\mathbf{z}}^{(k+1)} = \widetilde{\mathbf{z}}^{(k)} + a_{\text{P}}^{(k)} \mathbf{p}_{\text{P}}^{(k)}$.
19 $\quad\quad$ Update $\mathbf{r}_{\text{P}}^{(k+1)} = \mathbf{r}_{\text{P}}^{(k)} + a_{\text{P}}^{(k)} \mathbf{v}^{(k)}$, and update $\mathbf{u}_{\text{P}}^{(k+1)}$ by solving (4.3).
20 $\quad\quad$ Update $\beta_{\text{P}}^{(k+1)} = \frac{\langle \mathbf{r}_{\text{P}}^{(k+1)}, \mathbf{u}_{\text{P}}^{(k)} \rangle}{\langle \mathbf{r}_{\text{P}}^{(k)}, \mathbf{r}_{\text{P}}^{(k)} \rangle}$, and update $\mathbf{p}_{\text{P}}^{(k+1)} = -\mathbf{u}_{\text{P}}^{(k+1)} + \beta_{\text{P}}^{(k+1)} \mathbf{p}_{\text{P}}^{(k)}$.
21 $\quad\quad$ Update $\mathbf{v}_{\text{P}}^{(k+1)} = \left( \frac{\mathbf{R}^\top \mathbf{X}^\top \mathbf{X} \mathbf{R}}{n} + \lambda \mathbf{I}_p \right) \mathbf{p}_{\text{P}}^{(t+1)}$, and update $k = k + 1$.
22 $\quad$ **end**
23 $\quad$ Update $\mathbf{u}_{\text{D}}^{(t+1)} = \mathbf{r}_{\text{D}}^{(t+1)} - \mathbf{X} \mathbf{R} \widetilde{\mathbf{z}}_{\text{D}}^{(k+1)}$, and update $\beta_{\text{D}}^{(t+1)} = \frac{\langle \mathbf{r}_{\text{D}}^{(t+1)}, \mathbf{u}_{\text{D}}^{(t)} \rangle}{\langle \mathbf{r}_{\text{D}}^{(t)}, \mathbf{r}_{\text{D}}^{(t)} \rangle}$.
24 $\quad$ Update $\mathbf{p}_{\text{D}}^{(t+1)} = -\mathbf{u}_{\text{D}}^{(t+1)} + \beta_{\text{D}}^{(t+1)} \mathbf{p}_{\text{D}}^{(t)}$, and update $\mathbf{v}_{\text{D}}^{(t+1)} = \left( \frac{\mathbf{X} \mathbf{X}^\top}{n} + \lambda \mathbf{I}_n \right) \mathbf{p}_{\text{D}}^{(t+1)}$.
25 **end**

Also, the idea presented in Section can also be adopted to further reduce the number of calls to $m \times d$ scale sub-problems, which leads to the accelerated iterative primal-dual sketch (Acc-IPDS) algorithm (Summarized in Algorithm 7). In Acc-IPDS, we maintains the both vectors in primal space $\mathbf{u}_P, \mathbf{v}_P, \mathbf{r}_P$ and vectors in dual space $\mathbf{u}_D, \mathbf{v}_D, \mathbf{r}_D$, to make sure the updating directions for both primal variables and dual variables are conjugate with previous updating directions. Moreover, based on the residual vector $\mathbf{r}_P$, Acc-IPDS iteratively calls the solver to find solution of the following sketched linear system of scale $m \times d$:

$$\widehat{\mathbf{u}}_P^{(k)} = \arg\min_{\mathbf{u}} \mathbf{u}^\top \left( \frac{\mathbf{R}^\top \mathbf{X}^\top \mathbf{\Pi} \mathbf{\Pi}^\top \mathbf{X} \mathbf{R}}{2n} + \frac{\lambda}{2} \mathbf{I}_d \right) \mathbf{u} - \left\langle \mathbf{r}_P^{(k)}, \mathbf{u} \right\rangle. \qquad (4.3)$$

As we will show in the subsequent section, the number of calls for solving problem (4.3) only grows logarithmically with the inverse of approximation error.

## 5 Theoretical Analysis

In this section we present the theoretical analysis of various iterative sketching procedures, and defer the omitted proofs to Appendix A. First we will provide a unified analysis of Hessian sketch as dual random projection. The unified analysis basically follows the analysis of (Zhang et al., 2014) and (Pilanci and Wainwright, 2016), but simultaneously provide recovering guarantees for both *primal* and *dual* variables of interest. Then we move to the convergence analysis of the proposed accelerated IHS and IDRP algorithms, where we will show improved convergence speed over standard IHS and IDRP. Finally, we prove the iterative primal-dual sketch will converge to the optimum within iterations only grow logarithmically with the target approximation accuracy.

### 5.1 A unified analysis of Hessian Sketch and Dual Random Projection

In this section we provide a simple, unified analysis for the recovery performance of Hessian Sketch and Dual random projection. As in (Pilanci and Wainwright, 2016), we use the following notion of *Gaussian width* for any set $\mathcal{K} \subseteq \mathbb{R}^p$:

$$\mathbb{W}(\mathcal{K}) = \mathbb{E}\left[\sup_{\mathbf{w} \in \mathcal{K}} \langle \mathbf{w}, \mathbf{g} \rangle\right], \qquad (5.1)$$

where $\mathbf{g}$ is a random vector drawn from normal distribution $\mathcal{N}(\mathbf{0}, \mathbf{I}_p)$. Intuitively speaking, if the set $\mathcal{K}$ is restrictive to certain directions, then $\mathbb{W}(\mathcal{K})$ should be small as well (Vershynin, 2015). Given a set $\mathcal{K}$ and a random matrix $\mathbf{R} \in \mathbb{R}^{p \times d}$, the following quantities will be important for further analysis:

$$\rho_1(\mathcal{K}, \mathbf{R}) = \inf_{\mathbf{u} \in \mathcal{K} \cap \mathcal{S}^{p-1}} \mathbf{u}^\top \mathbf{R} \mathbf{R}^\top \mathbf{u} \quad \text{and} \quad \rho_2(\mathcal{K}, \mathbf{R}, \mathbf{v}) = \sup_{\mathbf{u} \in \mathcal{K} \cap \mathcal{S}^{p-1}} \left| \mathbf{u}^\top \left( \mathbf{R} \mathbf{R}^\top - \mathbf{I}_p \right) \mathbf{v} \right|,$$

where $\mathcal{S}^{p-1}$ is the $p$-dimensional unit-sphere. Firstly we could like the sketching matrix $\mathbf{R}$ to satisfy

$$\mathbb{E}\left[\mathbf{R}\mathbf{R}^\top\right] = \mathbf{I}_p,$$

and moreover, the matrix $\mathbf{R}\mathbf{R}^\top$ becomes closer to $\mathbf{I}_p$ as sketching dimension $d$ increases. Thus we would like to push $\rho_1(\mathcal{K}, \mathbf{R})$ to be close to 1, and $\rho_2(\mathcal{K}, \mathbf{R}, \mathbf{v})$ to be close to 0.

For the sake of simplicity, we just assume the random matrix $\mathbf{R}$ is sampled i.i.d. from some $\frac{1}{\sqrt{d}}$-sub-Gaussian distributions, this can be done by first sample a matrix $\widetilde{\mathbf{R}}$ where entries are sampled i.i.d. from 1-sub-Gaussian distribution, then perform the normalization as:

$$\mathbf{R} = \frac{\widetilde{\mathbf{R}}}{\sqrt{d}}.$$

The following lemma, from Pilanci and Wainwright (2015b), states how large the sketching dimension $d$ should be to make $\rho_1(\mathcal{K}, \mathbf{R}), \rho_2(\mathcal{K}, \mathbf{R}, \mathbf{v})$ to be close to 1 and 0, respectively.

**Lemma 5.1.** *When $\mathbf{R}$ is sampled i.i.d. from $1/\sqrt{d}$-sub-Gaussian distributions, then there exists universal constants $C_0$ such that, we have with probability at least $1 - \delta$, we have*

$$\rho_1(\mathcal{K}, \mathbf{R}) \geqslant 1 - C_0\sqrt{\frac{\mathbb{W}^2(\mathcal{K})}{d}} \log\left(\frac{1}{\delta}\right) \quad \text{and} \quad \rho_2(\mathcal{K}, \mathbf{R}, \mathbf{v}) \leqslant C_0\sqrt{\frac{\mathbb{W}^2(\mathcal{K})}{d}} \log\left(\frac{1}{\delta}\right).$$

For a set $\mathcal{K} \subseteq \mathbb{R}^p$, define the transformed set $\mathbf{X}^\top \mathcal{K}$ as where $\mathbf{X} \in \mathbb{R}^{n \times p}$

$$\mathbf{X}\mathcal{K} = \{\mathbf{u} \in \mathbb{R}^n | \mathbf{u} = \mathbf{X}\mathbf{v}, \mathbf{v} \in \mathcal{K}\}.$$

To present the main results about the unified analysis. Let's recall the main reductions in Hessian sketch and dual random projection. For Hessian sketch, we perform *sample reduction* with the transformation $\mathbf{X} \to \mathbf{\Pi}^\top \mathbf{X}$; for dual random projection, we perform *dimension reduction* with the transformation $\mathbf{X} \to \mathbf{X}\mathbf{R}$, where $\mathbf{\Pi} \in \mathbb{R}^{n \times m}, \mathbf{R} \in \mathbb{R}^{p \times d}$. Let $\widehat{\mathbf{w}}_{\text{HS}}$ be the approximated solution via Hessian sketch by solving (2.4), and the corresponding dual variables by the following transform

$$\widehat{\boldsymbol{\alpha}}_{\text{HS}} = \mathbf{y} - \mathbf{X}\widehat{\mathbf{w}}_{\text{HS}}.$$

Likewise, let $\widehat{\boldsymbol{\alpha}}_{\text{DRP}}$ and $\widehat{\mathbf{w}}_{\text{DRP}}$ be the approximated dual variables and primal variables obtained by dual random projection. The following theorem established the recovery bound for $\widehat{\boldsymbol{\alpha}}_{\text{HS}}, \widehat{\boldsymbol{\alpha}}_{\text{DRP}}$ and $\mathbf{w}_{\text{HS}}, \widehat{\mathbf{w}}_{\text{DRP}}$ simultaneously.

**Theorem 5.2.** *Suppose we perform Hessian sketch or dual random projection for problem (2.1) with sub-Gaussian sketching matrix $\mathbf{\Pi} \in \mathbb{R}^{n \times m}$ (for HS) or $\mathbf{R} \in \mathbb{R}^{p \times d}$ (for DRP). Then there exists universal constants $C_0$ such that with probability at least $1 - \delta$, the following approximation error bounds for HS or DRP holds:*
**For Hessian sketch:**

$$\|\widehat{\mathbf{w}}_{\text{HS}} - \mathbf{w}^*\|_{\mathbf{X}} \leqslant \frac{C_0\sqrt{\frac{\mathbb{W}^2(\mathbf{X}\mathbb{R}^p)}{m}} \log\left(\frac{1}{\delta}\right)}{1 - C_0\sqrt{\frac{\mathbb{W}^2(\mathbf{X}\mathbb{R}^p)}{m}} \log\left(\frac{1}{\delta}\right)} \|\mathbf{w}^*\|_{\mathbf{X}}, \tag{5.2}$$

$$\|\widehat{\boldsymbol{\alpha}}_{\text{HS}} - \boldsymbol{\alpha}^*\|_2 \leqslant \frac{\sqrt{n}C_0\sqrt{\frac{\mathbb{W}^2(\mathbf{X}\mathbb{R}^p)}{m}} \log\left(\frac{1}{\delta}\right)}{1 - C_0\sqrt{\frac{\mathbb{W}^2(\mathbf{X}\mathbb{R}^p)}{m}} \log\left(\frac{1}{\delta}\right)} \|\mathbf{w}^*\|_{\mathbf{X}}, \tag{5.3}$$

*For dual random projection:*

$$\|\widehat{\mathbf{w}}_{\text{DRP}} - \mathbf{w}^*\|_2 \leqslant \frac{C_0\sqrt{\frac{\mathbb{W}^2(\mathbf{X}^\top\mathbb{R}^n)}{d}}\log\left(\frac{1}{\delta}\right)}{1 - C_0\sqrt{\frac{\mathbb{W}^2(\mathbf{X}^\top\mathbb{R}^n)}{d}}\log\left(\frac{1}{\delta}\right)} \|\mathbf{w}^*\|_2, \tag{5.4}$$

$$\|\widehat{\boldsymbol{\alpha}}_{\text{DRP}} - \boldsymbol{\alpha}^*\|_{\mathbf{X}^\top} \leqslant \frac{C_0\sqrt{\frac{\mathbb{W}^2(\mathbf{X}^\top\mathbb{R}^n)}{d}}\log\left(\frac{1}{\delta}\right)}{1 - C_0\sqrt{\frac{\mathbb{W}^2(\mathbf{X}^\top\mathbb{R}^n)}{d}}\log\left(\frac{1}{\delta}\right)} \|\boldsymbol{\alpha}^*\|_{\mathbf{X}^\top}, \tag{5.5}$$

where the norm $\|\cdot\|_{\mathbf{X}}$ is defined as $\|\mathbf{w}\|_{\mathbf{X}} = \sqrt{\frac{\mathbf{w}^\top \mathbf{X}^\top \mathbf{X}\mathbf{w}}{n}}$.

**Remark**. We have the following remarks for Theorem 5.2.

- For general low-dimensional problems where $n \gg p$, $\mathbb{W}^2(\mathbf{X}\mathbb{R}^p) = p$, thus we have $\|\widehat{\mathbf{w}}_{\text{HS}} - \mathbf{w}^*\|_{\mathbf{X}} \lesssim \sqrt{\frac{p}{m}}\log\left(\frac{1}{\delta}\right)\|\mathbf{w}^*\|_{\mathbf{X}}$, which is the recovery bound proved in (Pilanci and Wainwright, 2016) (Proposition 1 in their paper).

- For high-dimensional problems when $p$ is large, $\mathbb{W}^2(\mathbf{X}^\top\mathbb{R}^n) = n$, thus we have $\|\widehat{\mathbf{w}}_{\text{DRP}} - \mathbf{w}^*\|_2 \lesssim \sqrt{\frac{n}{d}}\log\left(\frac{1}{\delta}\right)\|\mathbf{w}^*\|_2$. Moreover, when $\mathbf{X}$ is low-rank, i.e. $\text{rank}(\mathbf{X}) = r$ and $r \ll \min(n,p)$, we have $\mathbb{W}^2(\mathbf{X}^\top\mathbb{R}^n) = r$, thus we have $\|\widehat{\mathbf{w}}_{\text{DRP}} - \mathbf{w}^*\|_2 \lesssim \sqrt{\frac{r}{d}}\log\left(\frac{1}{\delta}\right)\|\mathbf{w}^*\|_2$, which is the recovery bound obtained in Theorem 1 of (Zhang et al., 2014), in fact the bound established in Theorem 5.2 improves Theorem 1 of (Zhang et al., 2014) by removing an additional $\sqrt{\log r}$ factor.

### 5.1.1 Analysis of IHS and DRP when X is approximately low-rank

In this section we provide recovery guarantees for the case when the data matrix $\mathbf{X}$ is *approximately* low rank. To make $\mathbf{X}$ can we well approximated by a rank $r$ matrix where $r \ll \min(n,p)$, we assume $\sigma_{r+1}$, the $r+1$-th singular value of $\mathbf{X}$, is small enough. Suppose $\mathbf{X}$ admits the following singular value decomposition:

$$\mathbf{X} := \mathbf{U}\boldsymbol{\Sigma}\mathbf{V}^\top = \mathbf{U}\boldsymbol{\Sigma}_r\mathbf{V}^\top + \mathbf{U}\boldsymbol{\Sigma}_{\bar{r}}\mathbf{V}^\top = \mathbf{X}_r + \mathbf{X}_{\bar{r}},$$

where $\bar{r}$ denotes the index $\{r+1, ..., \max(n,p)\}$. We also requires $\mathbf{w}^*$ can be well approximated by the a linear combination of top $r$ left singular vectors of $\mathbf{X}$, i.e. the remaining singular vectors are almost orthogonal with $\mathbf{w}^*$, depends on the method (Hessian sketch or dual random projection), we require the following notion of orthogonality holds for $\rho$ and $\varrho$ which are small:

$$\|\mathbf{X}_{\bar{r}}\mathbf{w}^*\|_2 \leqslant \rho\|\mathbf{X}\mathbf{w}^*\|_2, \qquad \|\mathbf{V}_{\bar{r}}^\top\mathbf{w}^*\|_2 \leqslant \varrho\|\mathbf{w}^*\|_2,$$

where $\mathbf{V}_{\bar{r}} \in \mathbb{R}^{p \times r}$ is the remaining right singular vectors of $\mathbf{X}$. Also, to simplify the results, let the entries of the sketching matrix $\boldsymbol{\Pi} \in \mathbb{R}^{m \times n}$ and $\mathbf{R} \in \mathbb{R}^{p \times d}$ are sampled i.i.d. from zero-mean Gaussian distributions, with variance $\frac{1}{m}$ and $\frac{1}{d}$, respectively. We have the following recovery bounds for Hessian sketch and dual random projection:

**Theorem 5.3.** *With probability at least $1 - \delta$, the following approximation error bounds for HS and DRP hold:*

**For Hessian sketch:**, *if*

$$m \geqslant \max\left(32(r+1), 4\log\left(\frac{2m}{\delta}\right), \frac{784\sigma_{r+1}^2}{9\lambda}\right)\log\left(\frac{n}{\delta}\right)$$

*then*

$$\|\widehat{\mathbf{w}}_{\mathrm{HS}} - \mathbf{w}^*\|_{\mathbf{X}} \leqslant 4\sqrt{\frac{1}{1-\epsilon_1} + \frac{\sigma_{r+1}^2}{\lambda n}} \cdot \sqrt{\frac{\epsilon_1^2 + \tau_1^2 \rho^2}{1-\epsilon_1} + \frac{\tau_1^2 \sigma_{r+1}^2 + \rho^2 v_1^2 \sigma_{r+1}^2}{\lambda n}} \|\mathbf{w}^*\|_{\mathbf{X}},$$

**For dual random projection:** *if*

$$d \geqslant \max\left(32(r+1), 4\log\left(\frac{2d}{\delta}\right), \frac{784p\sigma_{r+1}^2}{9\lambda n}\right)\log\left(\frac{p}{\delta}\right)$$

*then*

$$\|\widehat{\mathbf{w}}_{\mathrm{DRP}} - \mathbf{w}^*\|_2 \leqslant 4\sqrt{\frac{1}{1-\epsilon_2} + \frac{\sigma_{r+1}^2}{\lambda n}} \cdot \sqrt{\frac{\epsilon_2^2 + \tau_2^2 \varrho^2}{1-\epsilon_2} + \frac{\tau_2^2 \sigma_{r+1}^2 + \varrho^2 v_2^2 \sigma_{r+1}^2}{\lambda n}} \|\mathbf{w}^*\|_2, \tag{5.6}$$

*where $\epsilon_1, \epsilon_2, \tau_1, \tau_2, v_1, v_2$ are defined as*

$$\epsilon_1 = 2\sqrt{\frac{2(r+1)}{m}\log\frac{2r}{\delta}}, \qquad \epsilon_2 = 2\sqrt{\frac{2(r+1)}{d}\log\frac{2r}{\delta}},$$

$$\tau_1 = \frac{7}{3}\sqrt{\frac{2(n-r)}{m}\log\frac{n}{\delta}}, \qquad \tau_2 = \frac{7}{3}\sqrt{\frac{2(p-r)}{d}\log\frac{p}{\delta}},$$

$$v_1 = 2\sqrt{\frac{2(n-r+1)}{m}\log\frac{2(n-r)}{\delta}}, \qquad v_2 = 2\sqrt{\frac{2(p-r+1)}{d}\log\frac{2(p-r)}{\delta}}.$$

**Remark.** We make the following comments on Theorem 5.3:

- When $\sigma_{r+1} = 0$, i.e. $\mathbf{X}$ is exactly rank $r$, above results becomes

$$\|\widehat{\mathbf{w}}_{\mathrm{HS}} - \mathbf{w}^*\|_{\mathbf{X}} \lesssim \sqrt{\frac{r}{m}} \|\mathbf{w}^*\|_{\mathbf{X}}, \qquad \|\widehat{\mathbf{w}}_{\mathrm{DRP}} - \mathbf{w}^*\|_2 \lesssim \sqrt{\frac{r}{d}} \|\mathbf{w}^*\|_2 \tag{5.7}$$

  which reduces to the results in Theorem 5.2.

- We see that if we have $\sigma_{r+1}, \rho, \varrho$ sufficiently small in the following order, i.e. for Hessian sketch:

$$\sigma_{r+1} \lesssim \sqrt{\lambda}, \qquad \rho \lesssim \sqrt{\frac{r}{n}},$$

  for DRP:

$$\sigma_{r+1} \lesssim \sqrt{\frac{\lambda n}{p}}, \qquad \varrho \lesssim \sqrt{\frac{r}{p}},$$

  the guarantees (5.7) still hold.

## 5.2 Analysis of the accelerated IHS and IDRP methods

In this section we provide convergence analysis for the proposed Acc-IHS and Acc-IDRP approaches. As discussed before, since Acc-IHS and Acc-IDRP are preconditioned conjugate gradient methods on primal and dual problems, respectively, with a sketched Hessian as a preconditioner. By classical analysis of preconditioned conjugate gradient (Luenberger), we have the following convergence guarantees:

**Proposition 5.4.** *For Acc-IHS, we have*

$$\left\|\widehat{\mathbf{w}}_{\mathrm{HS}}^{(t)} - \mathbf{w}^*\right\|_{\mathbf{X}} \leqslant 2 \left( \frac{\sqrt{\kappa_{\mathrm{HS}}(\mathbf{X}, \mathbf{\Pi}, \lambda)} - 1}{\sqrt{\kappa_{\mathrm{HS}}(\mathbf{X}, \mathbf{\Pi}, \lambda)} + 1} \right)^t \|\mathbf{w}^*\|_{\mathbf{X}}, \tag{5.8}$$

*and Acc-IDRP, we have*

$$\left\|\widehat{\boldsymbol{\alpha}}_{\mathrm{DRP}}^{(t)} - \boldsymbol{\alpha}^*\right\|_{\mathbf{X}^\top} \leqslant 2 \left( \frac{\sqrt{\kappa_{\mathrm{DRP}}(\mathbf{X}, \mathbf{R}, \lambda)} - 1}{\sqrt{\kappa_{\mathrm{DRP}}(\mathbf{X}, \mathbf{R}, \lambda)} + 1} \right)^t \|\boldsymbol{\alpha}^*\|_{\mathbf{X}^\top}. \tag{5.9}$$

*where*

$$\kappa_{\mathrm{HS}}(\mathbf{X}, \mathbf{\Pi}, \lambda) = \kappa\left( \left( \frac{\mathbf{X}^\top \mathbf{\Pi} \mathbf{\Pi}^\top \mathbf{X}}{n} + \lambda \mathbf{I}_p \right)^{-1} \left( \frac{\mathbf{X}^\top \mathbf{X}}{n} + \lambda \mathbf{I}_p \right) \right),$$

$$\kappa_{\mathrm{DRP}}(\mathbf{X}, \mathbf{R}, \lambda) = \kappa\left( \left( \frac{\mathbf{X} \mathbf{R} \mathbf{R}^\top \mathbf{X}^\top}{n} + \lambda \mathbf{I}_n \right)^{-1} \left( \frac{\mathbf{X} \mathbf{X}^\top}{n} + \lambda \mathbf{I}_n \right) \right).$$

From Proposition 5.4, we know the convergence of Acc-IHS and Acc-IDRP heavily depends on the condition number $\kappa_{\mathrm{HS}}(\mathbf{X}, \mathbf{\Pi}, \lambda)$ and $\kappa_{\mathrm{DRP}}(\mathbf{X}, \mathbf{R}, \lambda)$. Thus the key of the rest of the analysis is to upper bound the condition numbers. To analyze the condition numbers, we make use of the following result in (Mendelson et al., 2007).

**Lemma 5.5.** *If the elements in $\mathbf{\Pi} \in \mathbb{R}^{n \times m}$ are i.i.d. sampled from a zero-mean $\frac{1}{m}$-sub-Gaussian distribution, then there exists universal constants $C_0$ such that, for any subset $\mathcal{K} \subseteq \mathbb{R}^n$, with probability at least $1 - \delta$, we have*

$$\sup_{\mathbf{u} \in \mathcal{K} \cap \mathcal{S}^{n-1}} \left| \mathbf{u}^\top \left( \mathbf{\Pi} \mathbf{\Pi}^\top - \mathbf{I}_n \right) \mathbf{u} \right| \leqslant C_0 \sqrt{\frac{\mathbb{W}^2(\mathcal{K})}{m}} \log\left(\frac{1}{\delta}\right).$$

Based on above lemma, we have the following bounds on the condition numbers $\kappa_{\mathrm{HS}}(\mathbf{X}, \mathbf{\Pi}, \lambda)$ and $\kappa_{\mathrm{DRP}}(\mathbf{X}, \mathbf{R}, \lambda)$:

**Theorem 5.6.** *If the sketching matrix $\mathbf{\Pi} \in \mathbb{R}^{n \times m}$ and $\mathbf{R} \in \mathbb{R}^{p \times d}$ are sampled from $\frac{1}{\sqrt{m}}$-sub-Gaussian and $\frac{1}{\sqrt{d}}$-sub-Gaussian distributions, repectively, then with probability at least $1 - \delta$, the* *following upper bounds hold:*

$$\kappa_{\mathrm{HS}}(\mathbf{X}, \mathbf{\Pi}, \lambda) \leqslant \frac{1}{1 - 2C_0\sqrt{\frac{\mathbb{W}^2(\mathbf{X}\mathbb{R}^p)}{m} \log\left(\frac{1}{\delta}\right)}}.$$

$$\kappa_{\mathrm{DRP}}(\mathbf{X}, \mathbf{R}, \lambda) \leqslant \frac{1}{1 - 2C_0\sqrt{\frac{\mathbb{W}^2(\mathbf{X}^\top\mathbb{R}^n)}{d} \log\left(\frac{1}{\delta}\right)}}.$$

With Theorem 5.6, we immediately obtain the following corollary which states the overall convergence for Acc-IHS and Acc-IDRP:

**Corollary 5.7.** *Suppose the sketching matrix $\mathbf{\Pi} \in \mathbb{R}^{n \times m}$ and $\mathbf{R} \in \mathbb{R}^{p \times d}$ are sub-Gaussian, if $t$, the number of iterations of Acc-IHS satisfies*

$$t \geqslant \left\lceil \left(\sqrt{\frac{1}{1 - 2C_0\sqrt{\frac{\mathbb{W}^2(\mathbf{X}\mathbb{R}^p)}{m} \log\left(\frac{1}{\delta}\right)}}}\right) \log\left(\frac{2\|\mathbf{w}^*\|_{\mathbf{X}}}{\epsilon}\right) \right\rceil$$

*then we have with probability at least $1 - \delta$,*

$$\left\|\widehat{\mathbf{w}}_{\mathrm{HS}}^{(t)} - \mathbf{w}^*\right\|_{\mathbf{X}} \leqslant \epsilon.$$

*If the number of iterations of Acc-IDRP satisfies*

$$t \geqslant \left\lceil \left(\sqrt{\frac{1}{1 - 2C_0\sqrt{\frac{\mathbb{W}^2(\mathbf{X}^\top\mathbb{R}^n)}{d} \log\left(\frac{1}{\delta}\right)}}}\right) \log\left(\frac{2\|\mathbf{w}^*\|_2}{\epsilon}\right) \right\rceil$$

*then we have with probability at least $1 - \delta$,*

$$\left\|\widehat{\mathbf{w}}_{\mathrm{DRP}}^{(t)} - \mathbf{w}^*\right\|_2 \leqslant \epsilon.$$

**Remark**. To compare the convergence rate of Acc-IHS, and Acc-IDRP with standard IHS (Pilanci and Wainwright, 2016) and IDRP (Zhang et al., 2014), we observe that

- For IHS, the number of iterations to reach $\epsilon$-accuracy (Corollary 1 in (Pilanci and Wainwright, 2016)) is $t \geqslant \left\lceil \left(\frac{1+\rho}{1-\rho}\right) \log\left(\frac{2\|\mathbf{w}^*\|_{\mathbf{X}}}{\epsilon}\right) \right\rceil$, where $\rho = C_0\sqrt{\frac{\mathbb{W}^2(\mathbf{X}\mathbb{R}^p)}{m} \log\left(\frac{1}{\delta}\right)}$. Acc-IHS reduces the number of iterations to $t \geqslant \left\lceil \left(\sqrt{\frac{1}{1-2\rho}}\right) \log\left(\frac{2\|\mathbf{w}^*\|_{\mathbf{X}}}{\epsilon}\right) \right\rceil$, which is significant when $\rho$ relatively large. Moreover, IHS requires $m \gtrsim \mathbb{W}^2(\mathbf{X}\mathbb{R}^p)$ to holds to converge, while Acc-IHS is always convergent.

- For IDRP, Theorem 7 in (Zhang et al., 2014) considered low-rank data, which requires IDRP to reach $\epsilon$-accuracy when $t \geqslant \left\lceil \left(\frac{1+\rho}{1-\rho}\right) \log\left(\frac{2\|\mathbf{w}^*\|_2}{\epsilon}\right) \right\rceil$, where $\rho = C_0\sqrt{\frac{r}{d} \log\left(\frac{r}{\delta}\right)}$. Acc-IDRP reduces the number of iterations to $t \geqslant \left\lceil \left(\sqrt{\frac{1}{1-2\rho}}\right) \log\left(\frac{2\|\mathbf{w}^*\|_2}{\epsilon}\right) \right\rceil$. Moreover, IDRP requires $d \gtrsim r \log r$ to holds to converge, while Acc-IDRP is always convergent.

## 5.3 Analysis for the primal-dual sketch methods

In this section, we provide runtime theoretical analysis for the proposed primal-dual sketch methods, where the sketched dual problem is not solved exactly, but apprxoimately solved via sketching the primal problem again. At outer loop iteration $t$, the standard analysis of iterative Hessian sketch ((Pilanci and Wainwright, 2016) and Theorem 5.2), we have the following lemma:

**Lemma 5.8.** *Let $\widehat{\mathbf{w}}_{\mathrm{HS}}^{(t+1)}$ be the iterate defined in Algorithm 1, then we have the following inequality:*

$$\left\|\widehat{\mathbf{w}}_{\mathrm{HS}}^{(t+1)} - \mathbf{w}^*\right\|_{\mathbf{X}} \leqslant \frac{C_0 \sqrt{\frac{\mathbb{W}^2(\mathbf{X}\mathbb{R}^p)}{m}} \log\left(\frac{1}{\delta}\right)}{1 - C_0 \sqrt{\frac{\mathbb{W}^2(\mathbf{X}\mathbb{R}^p)}{m}} \log\left(\frac{1}{\delta}\right)} \left\|\widehat{\mathbf{w}}_{\mathrm{HS}}^{(t)} - \mathbf{w}^*\right\|_{\mathbf{X}}.$$

However, in iterative primal-dual sketch, we don't have access to the exact minimizer $\widehat{\mathbf{w}}_{\mathrm{HS}}^{(t+1)}$, instead an *approximate* minimizer $\widetilde{\mathbf{w}}_{\mathrm{HS}}^{(t+1)}$ which is close to $\widehat{\mathbf{w}}_{\mathrm{HS}}^{(t+1)}$. The key is the analyze the iteration complexities of inner loops.

**Theorem 5.9.** *With probability at least $1 - \delta$, we have the following approximation error bound for $\widetilde{\mathbf{w}}_{\mathrm{HS}}^{(t+1)}$ in iterative primal-dual sketch:*

$$\left\|\widetilde{\mathbf{w}}_{\mathrm{HS}}^{(t+1)} - \mathbf{w}^*\right\|_{\mathbf{X}} \leqslant \left(\frac{C_0 \sqrt{\frac{\mathbb{W}^2(\mathbf{X}\mathbb{R}^p)}{m}} \log\left(\frac{1}{\delta}\right)}{1 - C_0 \sqrt{\frac{\mathbb{W}^2(\mathbf{X}\mathbb{R}^p)}{m}} \log\left(\frac{1}{\delta}\right)}\right)^t \|\mathbf{w}^*\|_{\mathbf{X}}$$

$$+ \frac{10\lambda_{\max}^2\left(\frac{\mathbf{X}^\top \mathbf{X}}{n}\right)}{\lambda^2} \left(\frac{C_0 \sqrt{\frac{\mathbb{W}^2(\mathbf{X}^\top \mathbb{R}^n)}{d}} \log\left(\frac{1}{\delta}\right)}{1 - C_0 \sqrt{\frac{\mathbb{W}^2(\mathbf{X}^\top \mathbb{R}^n)}{d}} \log\left(\frac{1}{\delta}\right)}\right)^k \|\mathbf{w}^*\|_2$$

With Theorem 5.9, we have the following iterative complexity for the proposed IPDS approach.

**Corollary 5.10.** *If the number of outer loops $t$ and number of inner loops $k$ in IPDS satisfying the following:*

$$t \geqslant \left\lceil \frac{1 + C_0 \sqrt{\frac{\mathbb{W}^2(\mathbf{X}\mathbb{R}^p)}{m}} \log\left(\frac{1}{\delta}\right)}{1 - C_0 \sqrt{\frac{\mathbb{W}^2(\mathbf{X}\mathbb{R}^p)}{m}} \log\left(\frac{1}{\delta}\right)} \right\rceil \log\left(\frac{4\|\mathbf{w}^*\|_{\mathbf{X}}}{\epsilon}\right)$$

$$k \geqslant \left\lceil \frac{1 + C_0 \sqrt{\frac{\mathbb{W}^2(\mathbf{X}^\top \mathbb{R}^n)}{d}} \log\left(\frac{1}{\delta}\right)}{1 - C_0 \sqrt{\frac{\mathbb{W}^2(\mathbf{X}^\top \mathbb{R}^n)}{d}} \log\left(\frac{1}{\delta}\right)} \right\rceil \log\left(\frac{40\lambda_{\max}^2\left(\frac{\mathbf{X}^\top \mathbf{X}}{n}\right)\|\mathbf{w}^*\|_2}{\lambda\epsilon}\right)$$

*Then with probability at least $1 - \delta$:*

$$\left\|\widetilde{\mathbf{w}}_{\mathrm{IPDS}}^{(t+1)} - \mathbf{w}^*\right\|_{\mathbf{X}} \leqslant \epsilon.$$

*Proof.* Apply Theorem 5.9 and substitute above inequalites for $t$ and $k$ we get $\left\|\widetilde{\mathbf{w}}_{\mathrm{IPDS}}^{(t+1)} - \mathbf{w}^*\right\|_{\mathbf{X}} \leqslant \frac{\epsilon}{2} + \frac{\epsilon}{2} = \epsilon$. □

**Remark.** Since the total number of sketched subproblem to solve in iterative primal-dual sketch is $tk$. To obtain $\epsilon$ approximation error, we have the total number of subproblems is

$$tk \lesssim \left\lceil \frac{1 + \sqrt{\frac{\mathbb{W}^2(\mathbf{X}\mathbb{R}^p)}{m}}}{1 - \sqrt{\frac{\mathbb{W}^2(\mathbf{X}\mathbb{R}^p)}{m}}} \cdot \frac{1 + \sqrt{\frac{\mathbb{W}^2(\mathbf{X}^\top\mathbb{R}^n)}{d}}}{1 - \sqrt{\frac{\mathbb{W}^2(\mathbf{X}^\top\mathbb{R}^n)}{d}}} \right\rceil \log^2\left(\frac{1}{\epsilon}\right).$$

Thus the iterative primal-dual sketch will be efficient when the Gaussian width of set $\mathbf{X}\mathbb{R}^p$ and $\mathbf{X}^\top\mathbb{R}^n$ is relatively small. For example, when $\text{rank}(\mathbf{X}) = r \ll \min(n,p)$, we can choose the sketching dimension in IPDS to be $m, d \gtrsim r$, and IPDS can return a solution with $\epsilon$-approximation error by just solving $\log^2\left(\frac{1}{\epsilon}\right)$ small scale subproblems of scale $r \times r$.

We next provide iteration complexity for the proposed Acc-IPDS algorithms as shown in Algorithm 7.

**Corollary 5.11.** *If the number of outer loops $t$ and number of inner loops $k$ in IPDS satisfying the following:*

$$t \geq \left\lceil \sqrt{\frac{1}{1 - 2C_0\sqrt{\frac{\mathbb{W}^2(\mathbf{X}\mathbb{R}^p)}{m}\log\left(\frac{1}{\delta}\right)}}} \right\rceil \log\left(\frac{4\|\mathbf{w}^*\|_\mathbf{X}}{\epsilon}\right)$$

$$k \geq \left\lceil \sqrt{\frac{1}{1 - 2C_0\sqrt{\frac{\mathbb{W}^2(\mathbf{X}^\top\mathbb{R}^n)}{d}\log\left(\frac{1}{\delta}\right)}}} \right\rceil \log\left(\frac{40\lambda_{\max}^2\left(\frac{\mathbf{X}^\top\mathbf{X}}{n}\right)\|\mathbf{w}^*\|_2}{\lambda\epsilon}\right)$$

*Then with probability at least $1 - \delta$:*

$$\left\|\widetilde{\mathbf{w}}_{\text{IPDS}}^{(t+1)} - \mathbf{w}^*\right\|_\mathbf{X} \leq \epsilon.$$

*Proof.* The proof is similar to the proof of Theorem 5.9 and then subsitute the lower bounds for $t$ and $k$ to obtain the result. □

### 5.4 Runtime comparison for large $n$, large $p$, and low-rank data

To solve problem (2.1), the runtime usually depends on several quantities: sample size $n$, problem dimension $p$ as well as the problem condition. To make the comparison simpler, we simply assume $\mathbf{X}$ is rank $r$, note that $r$ might be much smaller than $n, p$: $r \ll n, p$. For (2.1) generally the regularization parameter $\lambda$ is chosen at the order of $\mathcal{O}(1/\sqrt{n})$ to $\mathcal{O}(1/n)$ (Sridharan et al., 2009; Dhillon et al., 2013), here in favor of the iterative optimization algorithms we simply choose the large $\lambda$, i.e. of order $\mathcal{O}(1/\sqrt{n})$. For iterative optimization algorithms, the convergence usually depend on the smoothness of the problem. In (2.1), the smoothness parameter is $\lambda_{\max}\left(\frac{\mathbf{X}^\top\mathbf{X}}{n} + \lambda\mathbf{I}_p\right)$, which is often of the order $\mathcal{O}(p)$ (e.g. random sub-Gaussian design). To compare the runtime for solving (2.1), we consider the following methods:

- **Solving Linear System:** which solves the problem exactly using matrix inversion, which requires $\mathcal{O}(np^2 + p^3)$.

| Approach / Runtime | $\mathcal{O}(\cdot)$ | $\widetilde{\mathcal{O}}(\cdot)$ | Comment |
|---|---|---|---|
| Linear System | $np^2 + p^3$ | $np^2 + p^3$ | |
| LS with Low-rank SVD | $npr + r^3$ | $npr + r^3$ | |
| Gradient Descent | $\left(n^{1.5}p^2\right)\log\left(\frac{1}{\varepsilon}\right)$ | $n^{1.5}p^2$ | |
| Acc.Gradient Descent | $\left(n^{1.25}p^{1.5}\right)\log\left(\frac{1}{\varepsilon}\right)$ | $n^{1.25}p^{1.5}$ | |
| Coordinate Descent | $\left(n^{1.5}p\right)\log\left(\frac{1}{\varepsilon}\right)$ | $n^{1.5}p$ | |
| SVRG,SDCA,SAG | $\left(np + n^{0.5}p^2\right)\log\left(\frac{1}{\varepsilon}\right)$ | $np + n^{0.5}p^2$ | |
| Catalyst,APPA | $\left(np + n^{0.75}p^{1.5}\right)\log\left(\frac{1}{\varepsilon}\right)$ | $np + n^{0.75}p^{1.5}$ | |
| DSPDC | $npr + \left(nr + n^{0.75}p^{1.5}r\right)\log\left(\frac{1}{\varepsilon}\right)$ | $npr + n^{0.75}p^{1.5}r$ | |
| IHS + Catalyst | $np\log p + n^{0.25}p^{1.5}r\log^2\left(\frac{1}{\varepsilon}\right)$ | $np + n^{0.25}p^{1.5}r$ | Fast when $p \ll n$ |
| DRP + Exact | $np\log n + (nr^2 + r^3)\log\left(\frac{1}{\varepsilon}\right)$ | $np + nr^2 + r^3$ | Fast when $n \ll p$ |
| Iter.primal-dual sketch | $np\log p + (n + r^3)\log^2\left(\frac{1}{\varepsilon}\right)$ | $np + r^3$ | Fast when $r \ll \max(p,n)$ |

Table 2: Comparison of various approaches for solving large scale problems (2.1), the runtime depend on $n, p, r, \varepsilon$.

- **Linear System with Low-rank SVD**: if we have the factorization $\mathbf{X} = \mathbf{U}\mathbf{V}^\top$, where $\mathbf{U} \in \mathbb{R}^{n \times r}$, $\mathbf{V} \in \mathbb{R}^{p \times r}$. Then we can solve the matrix inversion efficiently with the Sherman-Morrison-Woodbury formula: $\left(\lambda \mathbf{I}_p + \frac{\mathbf{X}^\top \mathbf{X}}{n}\right)^{-1} = \frac{1}{\lambda}\mathbf{I}_p - \frac{1}{\lambda^2}\mathbf{V}\mathbf{U}^\top \mathbf{U}(\mathbf{I}_r + \mathbf{V}^\top \mathbf{V}\mathbf{U}^\top \mathbf{U})^{-1}\mathbf{V}^\top$. Which can be done in $\mathcal{O}(npr + r^3)$.

- **Gradient Descent:** standard analysis (Nesterov, 2013) shows gradient descent requires $\mathcal{O}\left(\left(\frac{L}{\lambda}\right)\log\left(\frac{1}{\varepsilon}\right)\right)$ iterations, with each iteration $\mathcal{O}(np)$, Since $L = \mathcal{O}(p), \lambda = \mathcal{O}(1/\sqrt{n})$. We have the overall runtime is $\mathcal{O}\left(\left(n^{1.5}p^2\right)\log\left(\frac{1}{\varepsilon}\right)\right)$.

- **Accelerated Gradient Descent** (Nesterov, 2013): which requires $\mathcal{O}\left(\sqrt{\left(\frac{L}{\lambda}\right)}\log\left(\frac{1}{\varepsilon}\right)\right)$ iterations, with each iteration $\mathcal{O}(np)$, Since $L = \mathcal{O}(p), \lambda = \mathcal{O}(1/\sqrt{n})$. We have the overall runtime is $\mathcal{O}\left(\left(n^{1.25}p^{1.5}\right)\log\left(\frac{1}{\varepsilon}\right)\right)$.

- **Randomized Coordinate Descent** (Nesterov, 2012): which requires $\mathcal{O}\left(p\left(\frac{1}{\lambda}\right)\log\left(\frac{1}{\varepsilon}\right)\right)$ iterations, with each iteration $\mathcal{O}(n)$, Since $\lambda = \mathcal{O}(1/\sqrt{n})$. We have the overall runtime is $\mathcal{O}\left(\left(n^{1.5}p\right)\log\left(\frac{1}{\varepsilon}\right)\right)$.

- **SVRG,SDCA,SAG** (Johnson and Zhang, 2013; Shalev-Shwartz and Zhang, 2013; Roux et al., 2012): which requires $\mathcal{O}\left(\left(n + \frac{L}{\lambda}\right)\log\left(\frac{1}{\varepsilon}\right)\right)$ iterations, with each iteration $\mathcal{O}(p)$. Since $L = \mathcal{O}(p), \lambda = \mathcal{O}(1/\sqrt{n})$. We have the overall runtime is $\mathcal{O}\left(\left(np + n^{0.5}p^2\right)\log\left(\frac{1}{\varepsilon}\right)\right)$.

- **Accelerated SVRG: Catalyst,APPA,SPDC,RPDG** (Lin et al., 2015; Frostig et al., 2015; Zhang and Lin, 2015; Lan and Zhou, 2015): which requires $\mathcal{O}\left(\left(n + \sqrt{n\frac{L}{\lambda}}\right)\log\left(\frac{1}{\varepsilon}\right)\right)$ iterations, with each iteration $\mathcal{O}(p)$. Since $L = \mathcal{O}(p), \lambda = \mathcal{O}(1/\sqrt{n})$. We have the overall runtime is $\mathcal{O}\left(\left(np + n^{0.75}p^{1.5}\right)\log\left(\frac{1}{\varepsilon}\right)\right)$.

- **DSPDC** (Yu et al., 2015): which requires $\mathcal{O}\left(\left(n + \sqrt{n\frac{L}{\lambda}p}\right)\log\left(\frac{1}{\varepsilon}\right)\right)$ iterations, with each iteration $\mathcal{O}(r)$. Since $L = \mathcal{O}(p), \lambda = \mathcal{O}(1/\sqrt{n})$. Also, to apply DSPDC, one should compute the low-rank factorization as a preprocessing step which takes $\mathcal{O}(npr)$. Thus we have the overall runtime is $\mathcal{O}\left(npr + \left(nr + n^{0.75}p^{0.5}r\right)\log\left(\frac{1}{\varepsilon}\right)\right)$.

- **Iterative Hessian Sketch + Accelerated SVRG (Pilanci and Wainwright, 2016):** compute sketched problem takes $\mathcal{O}(np\log p)$ (e.g. via fast Johnson-Lindenstrauss transforms (Ailon and Chazelle, 2009)), solve $\mathcal{O}\left(\log\left(\frac{1}{\varepsilon}\right)\right)$ sketched problems use accelerated SVRG type algorithm takes $\mathcal{O}\left(n^{0.25}p^{1.5}r\log\left(\frac{1}{\varepsilon}\right)\right)$. This leads to the overall runtime: $\mathcal{O}\left(np\log p + n^{0.25}p^{1.5}r\log^2\left(\frac{1}{\varepsilon}\right)\right)$.

- **DRP + Matrix inversion (Zhang et al., 2014):** compute sketched problem takes $\mathcal{O}(np\log n)$. Solve $\mathcal{O}\left(\log\left(\frac{1}{\varepsilon}\right)\right)$ reduced problem with matrix inversion takes $\mathcal{O}\left(nr^2 + r^3\right)$. This leads to the overall runtime: $\mathcal{O}\left(np\log n + (nr^2 + r^3)\log\left(\frac{1}{\varepsilon}\right)\right)$

- **Iterative Primal-Dual Sketch:** compute sketched problem takes $\mathcal{O}(np\log p)$. Solve $\mathcal{O}\left(\log^2\left(\frac{1}{\varepsilon}\right)\right)$, and solve reduced problem exactly takes $\mathcal{O}\left(n + r^3\right)$. Overall runtime: $\mathcal{O}\left(np\log p + (n + r^3)\log^2\left(\frac{1}{\varepsilon}\right)\right)$.

# 6 Experiments

In this section we present extensive comparisons for the proposed iterative sketching approaches on both simulated and real world data sets. We first demonstrate the improved convergence of the proposed Acc-IHS and Acc-IDRP algorithms on simulated data sets, and then show the proposed iterative primal-dual sketch procedure and its accelerated version could simultaneously reduce the sample size and dimensions of the problem, while still maintaining high approximation precision. Then we test these algorithms on some real world data sets.

## 6.1 Simulations for Acc-IHS and Acc-IDRP

We first examine the effectiveness of the proposed Acc-IHS and Acc-DRP algorithms on simulated data. The response variable $\{y_i\}_{i\in[n]}$ are drawn from the following linear model:

$$y_i = \langle \mathbf{x}_i, \boldsymbol{\beta}^* \rangle + \epsilon_i,$$

where the noise $\epsilon_{ji}$ is sampled from a standard normal distribution. The true model $\boldsymbol{\beta}^*$ is a $p$-dimensional vector where the entries are sampled i.i.d. from a uniform distribution in $[0, 1]$.

We first compare the proposed Acc-IHS with the standard IHS on some "big n", but relatively low-dimensional data. We generate $\{\mathbf{x}_i\}_{i\in[n]}$ from multivariate normal distribution with mean zero vector, and covariance matrix $\boldsymbol{\Sigma}$, which controls the condition number of the problem. We will varying $\boldsymbol{\Sigma}$ to see how it affects the performance of various methods. We set $\Sigma_{ij} = 0.5^{|i-j|}$ for the well-conditioned setting, and $\Sigma_{ij} = 0.5^{|i-j|/10}$ for the ill-conditioned setting. We fix the sample size $n = 10^5$ and varying the dimensions with $p = 50, 100, 300$. The results are shown in Figure 1, where for each problem setting, we test 3 different sketching dimensions (number inside parentheses in legend). We have the following observations:

- For both IHS and Acc-IHS, the larger the sketching dimension $m$, the faster the iterative converges to the optimum, which is consistent with the theory, as also observed in (Pilanci and Wainwright, 2016) and (Zhang et al., 2014) for IHS and IDRP algorithm.

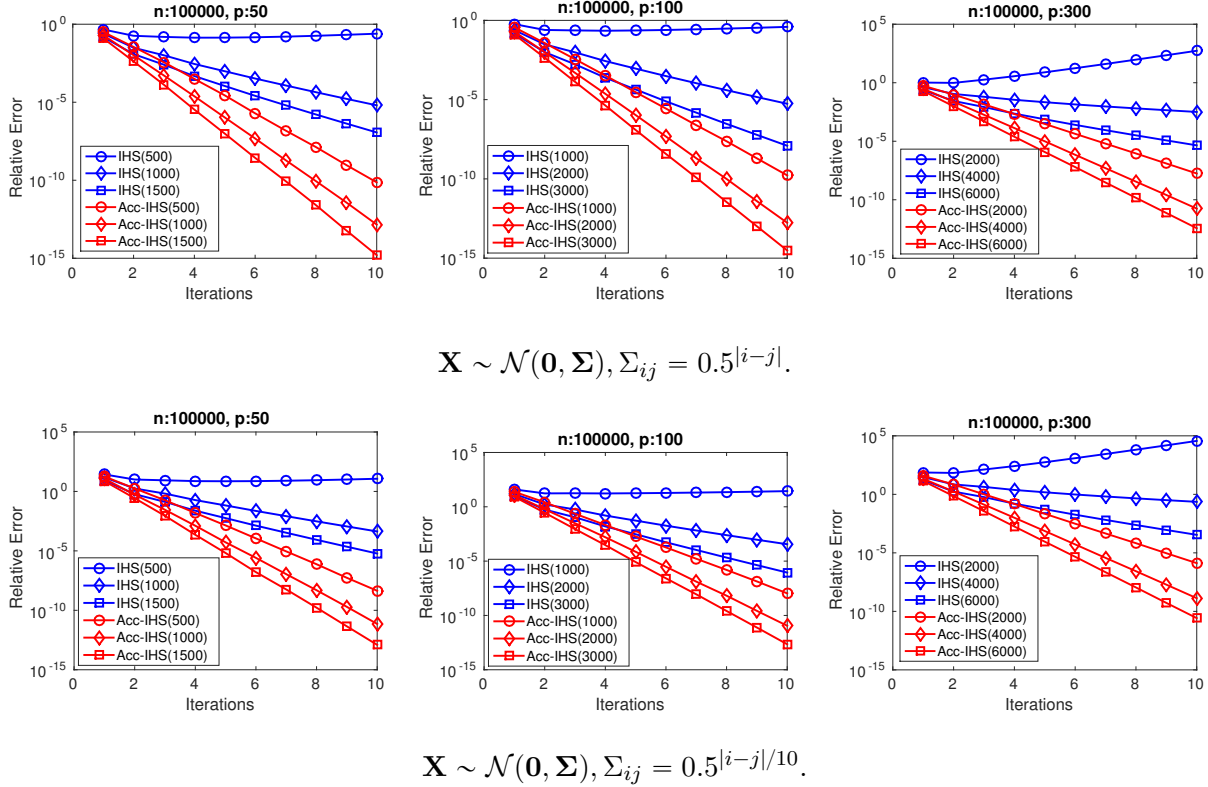

Figure 1: Comparion of IHS and Acc-IHS on various simulated datasets.

- When compared with IHS and Acc-IHS, we observed Acc-IHS converges significantly faster than IHS. Moreover, when the sketching dimension is small, IHS can diverge and go far away from the optimum, while Acc-IHS still converges.

- For all the cases we tested, Acc-IHS converges faster than IHS even when its sketching dimension is only 1/3 of the sketching dimension in IHS.

We then compare the proposed Acc-IDRP with the standard IDRP on high-dimensional, but relatively low-rank data. We generate $\{\mathbf{x}_i\}_{i\in[n]}$ from a low-rank factorization: $\mathbf{X} = \mathbf{U}\mathbf{V}^\top$, where the entries in $\mathbf{U} \in \mathbb{R}^{n\times r}, \mathbf{V} \in \mathbb{R}^{p\times r}$ are sampled i.i.d from standard normal distribution. We fix the sample size $n = 10^4$ and varying the dimensions with $p = 2000, 5000, 20000$, we also vary the rank $r = 20, 50$. The results are shown in Figure 2, where for each problem setting, we test 3 different sketching dimensions (number inside parentheses in legend). We have similar observations with the IHS case, i.e. Acc-IDRP always converges significantly faster than IDRP, even in the low sketching dimension case where IDRP diverge.

Above experiments validate the theoretical analysis which showed the accelerated procedures for IHS and IDRP could significantly boost the convergence of their standard counterpart. Since the computational cost per iteration of the standard iterative sketching techniques and their accelerated version is almost the same, thus Acc-IHS and Acc-IDRP will be useful techniques for iterative sketching with faster convergence speed.

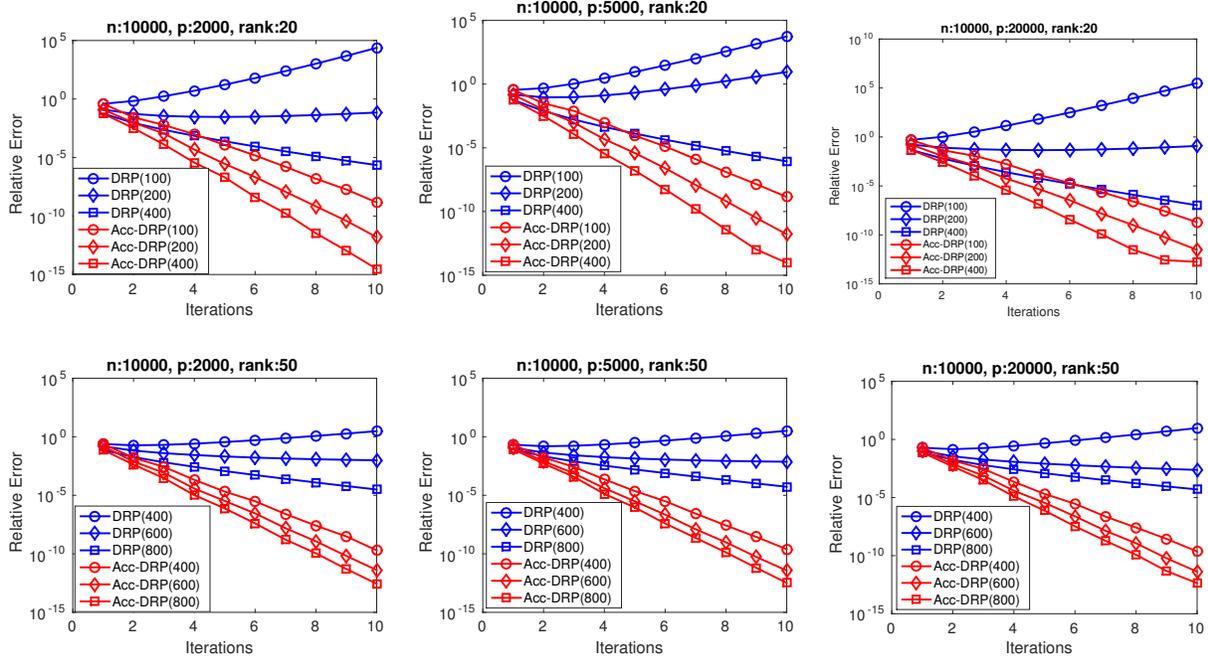

Figure 2: Comparion of IDRP and Acc-IDRP on various simulated datasets.

## 6.2 Simulations for IPDS and Acc-IPDS

In this section we demonstrate how iterative primal-dual sketch and its accelerated version works for simulated data. We generated the data using the same procedure as the simulation for Acc-DRP, where we fix the low-rank data with rank 10, and varing the original sample size $n$ and dimension $p$. For primal-dual sketching, we reduce the sample size to $m$, and dimension to $d$, where $m \ll n, d \ll p$. We also compare with standard IHS and IDRP, where for IHS we only perform sample reduction from $n$ to $m$, for IDRP only data dimension is reduced from $p$ to $d$. Thus the for subproblem size for IPDS (and Acc-IPDS), IHS, IDRP, are $m \times d$, $m \times p$, $n \times d$, respectively. For IPDS and Acc-IPDS, we terminate the inner loop when the $\ell_\infty$ distance between two inner iterations are less than $10^{-10}$. The results are shown in Figure 3, where the sketched dimension $(m, d)$ is shown in legend. We have the following observations:

- Though simultaneously reduce the sample size and data dimension, IPDS and Acc-IPDS are able to recover the optimum to a very high precision. However, they requires generally more iterations to reach certain approximation level compared with IHS and IDRP, where at each iteration we need to solve a substantial larger scale subproblem. Therefore, primal-dual sketching approach still enjoy more computational advantages. For example, on problem of scale $(n, p) = (10000, 20000)$, IHS and IDRP need to solve 5 sub-problems of scale $(m, p) = (500, 20000)$ and $(n, d) = (10000, 500)$, respectively, while for Acc-IPDS, it is only required to solve 35 sub-problems of scale $(m, d) = (500, 500)$ to obtain the same approximation accuracy.

- Acc-IPDS converges significantly faster than IPDS, which again verified the effectiveness of

Table 3: List of real-world data sets used in the experiments.

| Name | #Instances | #Features |
|---|---|---|
| connect4 | 67,557 | 126 |
| slice | 53,500 | 385 |
| year | 51,630 | 90 |
| colon-cancer | 62 | 2,000 |
| duke breast-cancer | 44 | 7,129 |
| leukemia | 72 | 7,129 |
| cifar | 4,047 | 3,072 |
| gisette | 6,000 | 5,000 |
| sector | 6,412 | 15,000 |

the proposed acceleration procedure for these sketching techniques.

### 6.3 Experiments on real datasets

We also conduct experiments on some real-world data sets where the statistics of them are summarized in Table 3. Among all the data sets, the first 3 are cases where sample size is significantly larger than the data dimension, where we used to compare the IHS and Acc-IHS algorithms; the middle 3 data sets are high-dimensional datasets with small sample size, where we compare to DRP and Acc-DRP procedures; the last 3 datasets are cases where sample size and data dimension are both relatively large, which is suitable for iterative primal-dual sketching methods. For the last 3 data sets we found that standard IHS and DRP often fails (unless with very large sketching dimension), thus we compared with Acc-IHS and Acc-DRP. We follow the same experimental setup with the simulation study, and the convergence plots are summarized in Figure 4. We have the following observations:

- Acc-IHS and Acc-DRP converges significantly faster than IHS and DRP, respectively, where similar observation is drawn from simulation studies.

- For the last 3 data sets where $n$ and $p$ are both large, and the data is not exactly low-rank: IHS, DRP and IPDS often diverge because of the requirement of the sketching dimension to ensure convergence is high, while the accelerated versions still converges to the optimum. It is notable that the Acc-IPDS only requires solving several least squares problems where both sample size and dimension are relatively small.

## 7 Conclusion and Discussion

In this paper, we focused on sketching techniques for solving large-scale $\ell_2$ regularized least square problems, we established the equivalence between the recently proposed two emerging techniques (Hessian sketch and dual random projection) from a primal-dual point of view, we also proposed accelerated methods for IHS and IDRP, from the preconditioned optimization perspective, and by combining the primal and dual sketching technique, we proposed a novel iterative primal-dual sketching approach which substantially reduced the computational cost in solving sketched subproblems.

The proposed approach can be extended to solving more general problems, for example, by sketching the Newton step in second-order optimization methods, as done in the "Newton Sketch" paper (Pilanci and Wainwright, 2015a), we will be able to solve regularized risk minimization problem with self-concordant losses, it will be interesting to examine its empirical performance compared with existing approaches. More generally, Hessian sketch and dual random projection are desingned for solving convex problems, it will be interesting to extend them for some structured non-convex problems, e.g. principle component analysis. Last by not least, it will be interesting to apply iterative sketching techniques for distributed optimization and learning problems (Heinze et al., 2014, 2015).

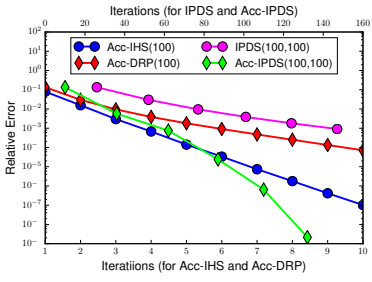
$(n,p) = (1000, 1000)$

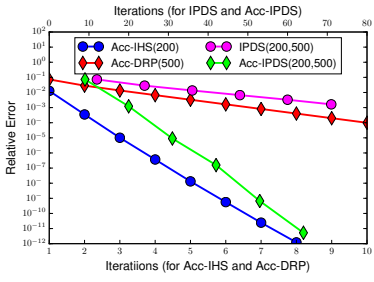
$(n,p) = (2000, 10000)$

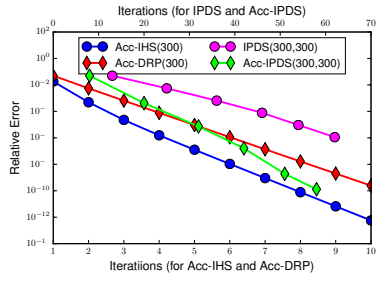
$(n,p) = (5000, 5000)$

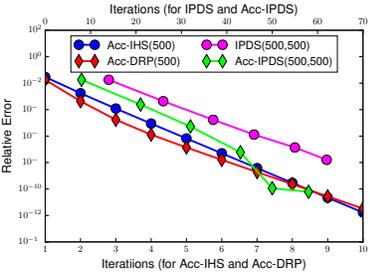
$(n,p) = (5000, 10000)$

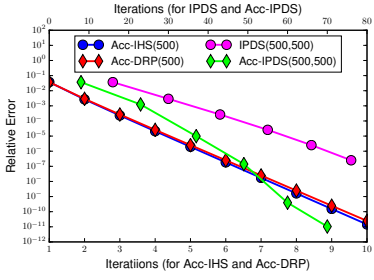
$(n,p) = (5000, 20000)$

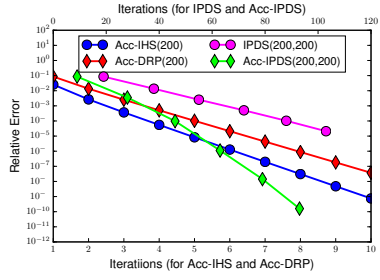
$(n,p) = (10000, 1000)$

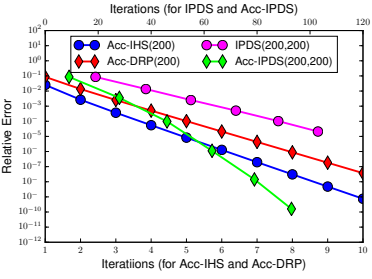
$(n,p) = (10000, 1000)$

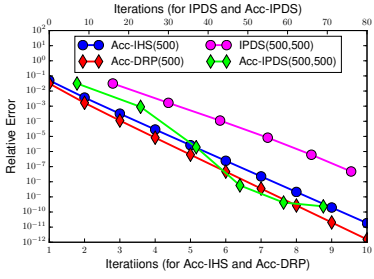
$(n,p) = (10000, 20000)$

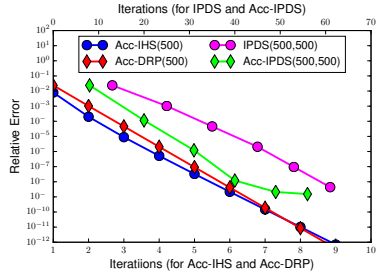
$(n,p) = (20000, 2000)$

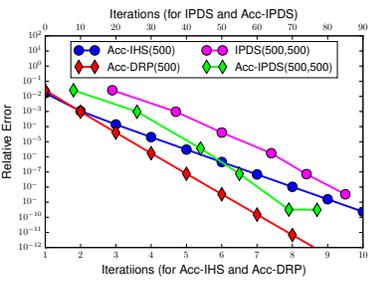
$(n,p) = (20000, 5000)$

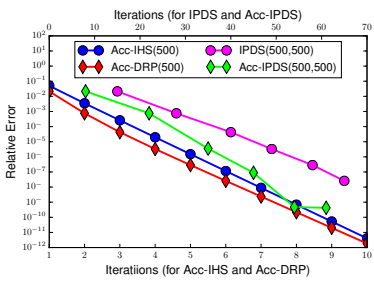
$(n,p) = (20000, 10000)$

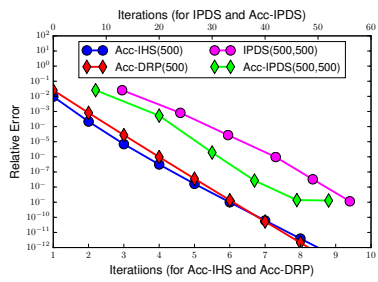
$(n,p) = (20000, 20000)$

Figure 3: Comparion of IPDS and Acc-IPDS versus with IHS and DRP various simulated datasets.

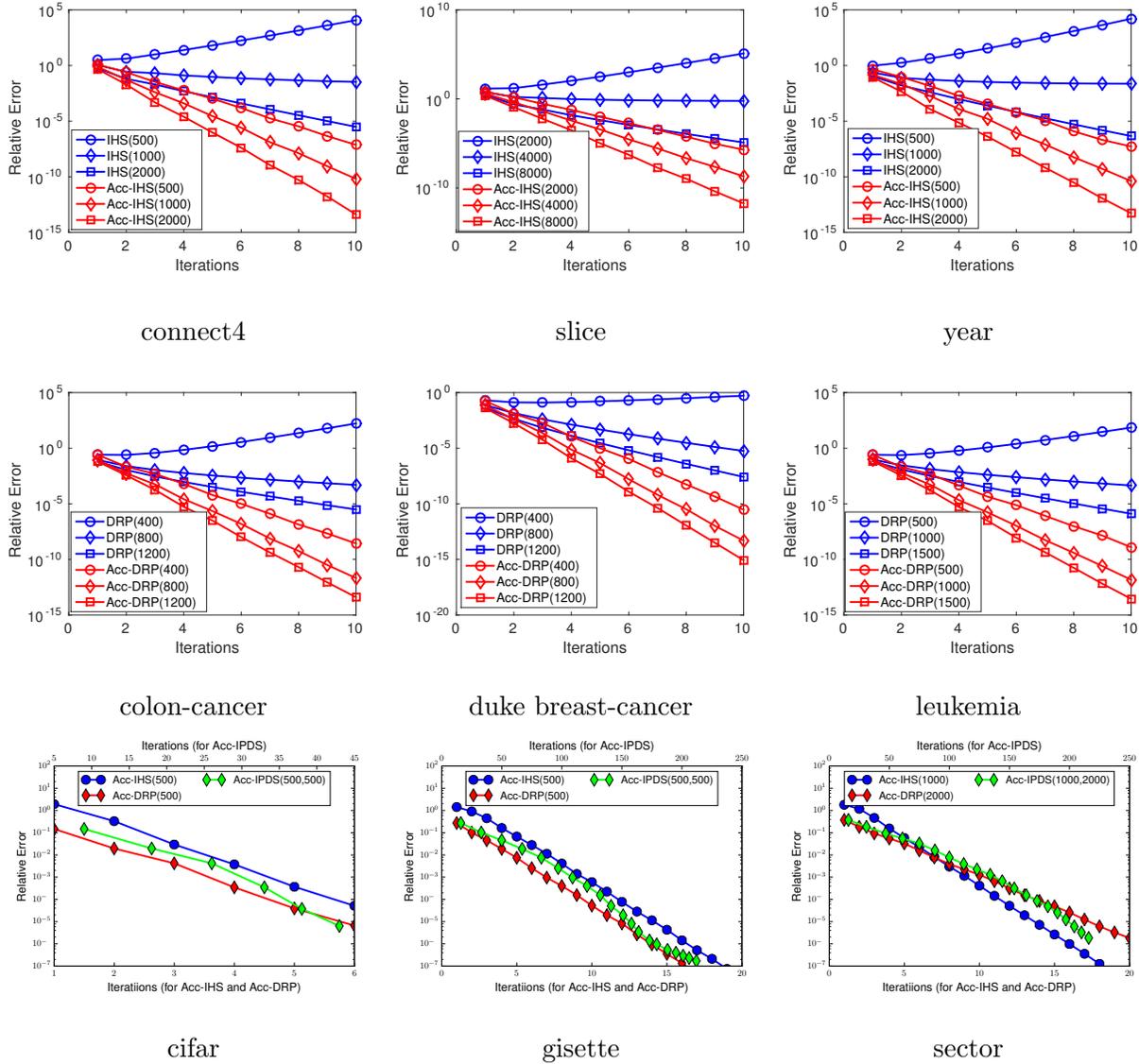

Figure 4: Comparion of various iterative sketching approaches on real-world datasets, Top row: Acc-IHS versus IHS, middle row: Acc-DRP versus DRP, bottom row: Acc-IPDS versus Acc-IHS and Acc-DRP.

# A  Appendix

The appendix contains proofs of theorems stated in the main paper.

## A.1  Proof of Theorem 5.2

*Proof.* Based on the optimality condition for $\mathbf{w}^*$ and $\widehat{\mathbf{w}}_{\text{HS}}$:

$$\left(\frac{\mathbf{X}^\top \mathbf{X}}{n} + \lambda \mathbf{I}_p\right) \mathbf{w}^* = \frac{\mathbf{X}^\top \mathbf{y}}{n}.$$

$$\left(\frac{\mathbf{X}^\top \mathbf{\Pi}\mathbf{\Pi}^\top \mathbf{X}}{n} + \lambda \mathbf{I}_p\right) \widehat{\mathbf{w}}_{\text{HS}} = \frac{\mathbf{X}^\top \mathbf{y}}{n}.$$

Thus we have

$$\left(\frac{\mathbf{X}^\top \mathbf{X}}{n} + \lambda \mathbf{I}_p\right) \mathbf{w}^* - \left(\frac{\mathbf{X}^\top \mathbf{\Pi}\mathbf{\Pi}^\top \mathbf{X}}{n} + \lambda \mathbf{I}_p\right) \widehat{\mathbf{w}}_{\text{HS}} = \mathbf{0}.$$

So

$$\left\langle \left(\frac{\mathbf{X}^\top \mathbf{X}}{n} + \lambda \mathbf{I}_p\right) \mathbf{w}^* - \left(\frac{\mathbf{X}^\top \mathbf{\Pi}\mathbf{\Pi}^\top \mathbf{X}}{n} + \lambda \mathbf{I}_p\right) \widehat{\mathbf{w}}_{\text{HS}}, \mathbf{w}^* - \widehat{\mathbf{w}}_{\text{HS}} \right\rangle = 0.$$

By adding and subtracting $\left\langle \mathbf{w}^* - \widehat{\mathbf{w}}_{\text{HS}}, \left(\frac{\mathbf{X}^\top \mathbf{\Pi}\mathbf{\Pi}^\top \mathbf{X}}{n} + \lambda \mathbf{I}_p\right) \mathbf{w}^* \right\rangle$ we have

$$\left\langle \left(\frac{\mathbf{X}^\top \mathbf{\Pi}\mathbf{\Pi}^\top \mathbf{X}}{n} - \frac{\mathbf{X}^\top \mathbf{X}}{n}\right) \mathbf{w}^*, \widehat{\mathbf{w}}_{\text{HS}} - \mathbf{w}^* \right\rangle = (\mathbf{w}^* - \widehat{\mathbf{w}}_{\text{HS}})^\top \left(\frac{\mathbf{X}^\top \mathbf{\Pi}\mathbf{\Pi}^\top \mathbf{X}}{n} + \lambda \mathbf{I}_p\right) (\mathbf{w}^* - \widehat{\mathbf{w}}_{\text{HS}})$$

For the term on right hand side, we have

$$\begin{aligned}(\mathbf{w}^* - \widehat{\mathbf{w}}_{\text{HS}})^\top \left(\frac{\mathbf{X}^\top \mathbf{\Pi}\mathbf{\Pi}^\top \mathbf{X}}{n} + \lambda \mathbf{I}_p\right) (\mathbf{w}^* - \widehat{\mathbf{w}}_{\text{HS}}) =& (\mathbf{w}^* - \widehat{\mathbf{w}}_{\text{HS}})^\top \left(\frac{\mathbf{X}^\top \mathbf{\Pi}\mathbf{\Pi}^\top \mathbf{X}}{n}\right) (\mathbf{w}^* - \widehat{\mathbf{w}}_{\text{HS}}) \\ & + \lambda \|\mathbf{w}^* - \widehat{\mathbf{w}}_{\text{HS}}\|_2^2 \\ \geqslant & \rho_1(\mathbf{X}\mathbb{R}^p, \mathbf{\Pi}) \|\mathbf{w}^* - \widehat{\mathbf{w}}_{\text{HS}}\|_{\mathbf{X}}^2.\end{aligned} \quad (A.1)$$

For the term on the left hand side, we have

$$\begin{aligned}\left\langle \left(\frac{\mathbf{X}^\top \mathbf{\Pi}\mathbf{\Pi}^\top \mathbf{X}}{n} - \frac{\mathbf{X}^\top \mathbf{X}}{n}\right) \mathbf{w}^*, \widehat{\mathbf{w}}_{\text{HS}} - \mathbf{w}^* \right\rangle =& \left\langle (\mathbf{\Pi}\mathbf{\Pi}^\top - \mathbf{I}_n) \frac{\mathbf{X}\mathbf{w}^*}{\sqrt{n}}, \frac{\mathbf{X}}{\sqrt{n}}(\widehat{\mathbf{w}}_{\text{HS}} - \mathbf{w}^*) \right\rangle \\ \leqslant & \rho_2(\mathbf{X}\mathbb{R}^p, \mathbf{\Pi}, \mathbf{w}^*) \|\mathbf{w}^*\|_{\mathbf{X}} \|\widehat{\mathbf{w}}_{\text{HS}} - \mathbf{w}^*\|_{\mathbf{X}}.\end{aligned} \quad (A.2)$$

Combining (A.1) and (A.2) we have

$$\|\widehat{\mathbf{w}}_{\text{HS}} - \mathbf{w}^*\|_{\mathbf{X}} \leqslant \frac{\rho_2(\mathbf{X}\mathbb{R}^p, \mathbf{\Pi}, \mathbf{w}^*)}{\rho_1(\mathbf{X}\mathbb{R}^p, \mathbf{\Pi})} \|\mathbf{w}^*\|_{\mathbf{X}}$$

For the recovery of dual variables, we have

$$\|\widehat{\boldsymbol{\alpha}}_{\text{HS}} - \boldsymbol{\alpha}^*\|_2 = \|\mathbf{y} - \mathbf{X}\widehat{\mathbf{w}}_{\text{HS}} - (\mathbf{y} - \mathbf{X}\mathbf{w}^*)\|_2 = \sqrt{n} \|\widehat{\mathbf{w}}_{\text{HS}} - \mathbf{w}^*\|_{\mathbf{X}} \leqslant \sqrt{n}\frac{\rho_2(\mathbf{X}\mathbb{R}^p, \mathbf{\Pi}, \mathbf{w}^*)}{\rho_1(\mathbf{X}\mathbb{R}^p, \mathbf{\Pi})} \|\mathbf{w}^*\|_{\mathbf{X}}.$$

For dual random projection, the proof is mostly analogous to the proof for Hessian sketch. Based on the optimality condition for $\boldsymbol{\alpha}^*$ and $\widehat{\boldsymbol{\alpha}}_{\text{DRP}}$:

$$\left(\frac{\mathbf{XX}^\top}{n} + \lambda \mathbf{I}_n\right)\boldsymbol{\alpha}^* = \lambda \mathbf{y},$$

$$\left(\frac{\mathbf{XRR}^\top\mathbf{X}^\top}{n} + \lambda \mathbf{I}_n\right)\widehat{\boldsymbol{\alpha}}_{\text{DRP}} = \lambda \mathbf{y}.$$

Thus we have

$$\left(\frac{\mathbf{XX}^\top}{n} + \lambda \mathbf{I}_n\right)\boldsymbol{\alpha}^* - \left(\frac{\mathbf{XRR}^\top\mathbf{X}^\top}{n} + \lambda \mathbf{I}_n\right)\widehat{\boldsymbol{\alpha}}_{\text{DRP}} = \mathbf{0}.$$

So

$$\left\langle \left(\frac{\mathbf{XX}^\top}{n} + \lambda \mathbf{I}_n\right)\boldsymbol{\alpha}^* - \left(\frac{\mathbf{XRR}^\top\mathbf{X}^\top}{n} + \lambda \mathbf{I}_n\right)\widehat{\boldsymbol{\alpha}}_{\text{DRP}}, \boldsymbol{\alpha}^* - \widehat{\boldsymbol{\alpha}}_{\text{DRP}}\right\rangle = 0.$$

By some algebraic manipulations we have

$$\left\langle \left(\frac{\mathbf{XRR}^\top\mathbf{X}^\top}{n} - \frac{\mathbf{XX}^\top}{n}\right)\boldsymbol{\alpha}^*, \widehat{\boldsymbol{\alpha}}_{\text{DRP}} - \boldsymbol{\alpha}^*\right\rangle = (\boldsymbol{\alpha}^* - \widehat{\boldsymbol{\alpha}}_{\text{DRP}})^\top \left(\frac{\mathbf{XRR}^\top\mathbf{X}^\top}{n} + \lambda \mathbf{I}_n\right)(\boldsymbol{\alpha}^* - \widehat{\boldsymbol{\alpha}}_{\text{DRP}}).$$

For the term on right hand side, we have

$$\begin{aligned}(\boldsymbol{\alpha}^* - \widehat{\boldsymbol{\alpha}}_{\text{DRP}})^\top \left(\frac{\mathbf{XRR}^\top\mathbf{X}^\top}{n} + \lambda \mathbf{I}_n\right)(\boldsymbol{\alpha}^* - \widehat{\boldsymbol{\alpha}}_{\text{DRP}}) =& (\boldsymbol{\alpha}^* - \widehat{\boldsymbol{\alpha}}_{\text{DRP}})^\top \left(\frac{\mathbf{XRR}^\top\mathbf{X}^\top}{n}\right)(\boldsymbol{\alpha}^* - \widehat{\boldsymbol{\alpha}}_{\text{DRP}}) \\ &+ \lambda \|\boldsymbol{\alpha}^* - \widehat{\boldsymbol{\alpha}}_{\text{DRP}}\|_2^2 \\ \geqslant & \rho_1(\mathbf{X}^\top \mathbb{R}^n, \mathbf{R}) \|\boldsymbol{\alpha}^* - \widehat{\boldsymbol{\alpha}}_{\text{DRP}}\|_{\mathbf{X}^\top}^2. \end{aligned} \quad (\text{A.3})$$

For the term on the left hand side, we have

$$\begin{aligned}\left\langle \left(\frac{\mathbf{XRR}^\top\mathbf{X}^\top}{n} - \frac{\mathbf{XX}^\top}{n}\right)\boldsymbol{\alpha}^*, \widehat{\boldsymbol{\alpha}}_{\text{DRP}} - \boldsymbol{\alpha}^*\right\rangle =& \left\langle (\mathbf{RR}^\top - \mathbf{I}_p)\frac{\mathbf{X}^\top\boldsymbol{\alpha}^*}{\sqrt{n}}, \frac{\mathbf{X}^\top}{\sqrt{n}}(\widehat{\boldsymbol{\alpha}}_{\text{DRP}} - \boldsymbol{\alpha}^*)\right\rangle \\ \leqslant & \rho_2(\mathbf{X}^\top\mathbb{R}^n, \mathbf{R}, \boldsymbol{\alpha}^*)\|\boldsymbol{\alpha}^*\|_{\mathbf{X}^\top}\|\widehat{\boldsymbol{\alpha}}_{\text{DRP}} - \boldsymbol{\alpha}^*\|_{\mathbf{X}^\top}. \end{aligned} \quad (\text{A.4})$$

Combining (A.3) and (A.4) we have

$$\|\widehat{\boldsymbol{\alpha}}_{\text{DRP}} - \boldsymbol{\alpha}^*\|_{\mathbf{X}^\top} \leqslant \frac{\rho_2(\mathbf{X}^\top\mathbb{R}^n, \mathbf{R}, \boldsymbol{\alpha}^*)}{\rho_1(\mathbf{X}^\top\mathbb{R}^n, \mathbf{R})}\|\boldsymbol{\alpha}^*\|_{\mathbf{X}^\top}.$$

For the recovery of primal variables, we have

$$\begin{aligned}\|\widehat{\mathbf{w}}_{\text{DRP}} - \mathbf{w}^*\|_2 =& \frac{1}{\lambda\sqrt{n}}\|\widehat{\boldsymbol{\alpha}}_{\text{DRP}} - \boldsymbol{\alpha}^*\|_{\mathbf{X}^\top} \leqslant \frac{1}{\lambda\sqrt{n}}\frac{\rho_2(\mathbf{X}^\top\mathbb{R}^n, \mathbf{R}, \boldsymbol{\alpha}^*)}{\rho_1(\mathbf{X}^\top\mathbb{R}^n, \mathbf{R})}\|\boldsymbol{\alpha}^*\|_{\mathbf{X}^\top} \\ =& \frac{\rho_2(\mathbf{X}^\top\mathbb{R}^n, \mathbf{R}, \boldsymbol{\alpha}^*)}{\rho_1(\mathbf{X}^\top\mathbb{R}^n, \mathbf{R})}\|\mathbf{w}^*\|_2. \end{aligned}$$

Then applying Lemma 5.1 we conclude the proof. $\square$

## A.2 Proof of Theorem 5.3

*Proof.* We only prove the results for Hessian sketch here as the proof for dual random projection is analogous and similar results already appeared in (Zhang et al., 2014). We will make usage of the following concentration results of Gaussian random matrix:

**Lemma A.1.** *(Lemma 3 in (Zhang et al., 2014)) Let $\mathbf{B} \in \mathbb{R}^{r \times m}$ is a random matrix with entries sampled i.i.d. from zero-mean Gaussian distribution with variance $\frac{1}{m}$, then with probability at least $1 - \delta$, the following inequality holds:*

$$\left\|\mathbf{B}\mathbf{B}^\top - \mathbf{I}_r\right\|_2 \leq 2\sqrt{\frac{2(r+1)}{m} \log \frac{2r}{\delta}} := \epsilon_1.$$

**Lemma A.2.** *(Theorem 3.2 in (Recht, 2011)) Let $\mathbf{B} \in \mathbb{R}^{r \times m}$, $\mathbf{A} \in \mathbb{R}^{(n-r) \times m}$ are random matrices with entries sampled i.i.d. from zero-mean Gaussian distribution with variance $\frac{1}{m}$, then with probability at least $1 - \delta$, the following inequality holds:*

$$\left\|\mathbf{A}\mathbf{B}^\top\right\|_2 \leq \frac{7}{3}\sqrt{\frac{2(n-r)}{m} \log \frac{n}{\delta}} := \tau_1.$$

We are ready to prove, let $\Delta \mathbf{w} = \mathbf{w}^* - \widehat{\mathbf{w}}_{\text{HS}}$, it is easy to see

$$\|\Delta \mathbf{w}\|_{\mathbf{X}}^2 = \left\|\frac{\mathbf{X}\Delta \mathbf{w}}{\sqrt{n}}\right\|_2^2 = \left\|\frac{(\mathbf{U}\boldsymbol{\Sigma}_r \mathbf{V}^\top + \mathbf{U}\boldsymbol{\Sigma}_{\bar{r}} \mathbf{V}^\top)\Delta \mathbf{w}}{\sqrt{n}}\right\|_2^2 = \left\|\frac{\boldsymbol{\Sigma}_r \mathbf{V}^\top \Delta \mathbf{w}}{\sqrt{n}}\right\|_2^2 + \left\|\frac{\boldsymbol{\Sigma}_{\bar{r}} \mathbf{V}^\top \Delta \mathbf{w}}{\sqrt{n}}\right\|_2^2$$

Consider the term $\Delta \mathbf{w}^\top \left(\frac{\mathbf{X}^\top \mathbf{\Pi}\mathbf{\Pi}^\top \mathbf{X}}{n} + \lambda \mathbf{I}_p\right) \Delta \mathbf{w}$, we have

$$\Delta \mathbf{w}^\top \left(\frac{\mathbf{X}^\top \mathbf{\Pi}\mathbf{\Pi}^\top \mathbf{X}}{n} + \lambda \mathbf{I}_p\right) \Delta \mathbf{w} \geq \Delta \mathbf{w}^\top \left(\frac{\mathbf{V}^\top \boldsymbol{\Sigma}_r \mathbf{U}^\top \mathbf{\Pi}\mathbf{\Pi}^\top \mathbf{U}\boldsymbol{\Sigma}_r \mathbf{V}^\top}{n}\right) \Delta \mathbf{w} + \lambda \|\Delta \mathbf{w}\|_2^2$$
$$+ 2\Delta \mathbf{w}^\top \left(\frac{\mathbf{V}^\top \boldsymbol{\Sigma}_{\bar{r}} \mathbf{U}^\top \mathbf{\Pi}\mathbf{\Pi}^\top \mathbf{U}\boldsymbol{\Sigma}_r \mathbf{V}^\top}{n}\right) \Delta \mathbf{w}.$$

Since

$$\Delta \mathbf{w}^\top \left(\frac{\mathbf{V}^\top \boldsymbol{\Sigma}_r \mathbf{U}^\top \mathbf{\Pi}\mathbf{\Pi}^\top \mathbf{U}\boldsymbol{\Sigma}_r \mathbf{V}^\top}{n}\right) \Delta \mathbf{w} \geq (1 - \epsilon_1) \left\|\frac{\boldsymbol{\Sigma}_r \mathbf{V}^\top \Delta \mathbf{w}}{\sqrt{n}}\right\|_2^2,$$

and

$$\lambda \|\Delta \mathbf{w}\|_2^2 \geq \frac{\lambda}{\sigma_{r+1}^2} \left\|\boldsymbol{\Sigma}_{\bar{r}} \mathbf{V}^\top \Delta \mathbf{w}\right\|_2^2 = \frac{\lambda n}{\sigma_{r+1}^2} \left\|\frac{\boldsymbol{\Sigma}_{\bar{r}} \mathbf{V}^\top \Delta \mathbf{w}}{\sqrt{n}}\right\|_2^2$$

where

$$2\Delta \mathbf{w}^\top \left(\frac{\mathbf{V}^\top \boldsymbol{\Sigma}_{\bar{r}} \mathbf{U}^\top \mathbf{\Pi}\mathbf{\Pi}^\top \mathbf{U}\boldsymbol{\Sigma}_r \mathbf{V}^\top}{n}\right) \Delta \mathbf{w} = 2\Delta \mathbf{w}^\top \left(\frac{\mathbf{V}^\top \boldsymbol{\Sigma}_{\bar{r}} \mathbf{U}_{\bar{r}}^\top \mathbf{\Pi}\mathbf{\Pi}^\top \mathbf{U}_r \boldsymbol{\Sigma}_r \mathbf{V}^\top}{n}\right) \Delta \mathbf{w}$$
$$\geq -\tau_1 \left\|\frac{\boldsymbol{\Sigma}_r \mathbf{V}^\top \Delta \mathbf{w}}{\sqrt{n}}\right\|_2 \left\|\frac{\boldsymbol{\Sigma}_{\bar{r}} \mathbf{V}^\top \Delta \mathbf{w}}{\sqrt{n}}\right\|_2$$

Combining above we have

$$\Delta\mathbf{w}^\top \left(\frac{\mathbf{X}^\top \mathbf{\Pi}\mathbf{\Pi}^\top \mathbf{X}}{n} + \lambda \mathbf{I}_p\right) \Delta\mathbf{w} \geq (1-\epsilon_1)\left\|\frac{\mathbf{\Sigma}_r \mathbf{V}^\top \Delta\mathbf{w}}{\sqrt{n}}\right\|_2^2 + \frac{\lambda n}{\sigma_{r+1}^2}\left\|\frac{\mathbf{\Sigma}_{\bar{r}} \mathbf{V}^\top \Delta\mathbf{w}}{\sqrt{n}}\right\|_2^2$$

$$- 2\tau_1 \left\|\frac{\mathbf{\Sigma}_r \mathbf{V}^\top \Delta\mathbf{w}}{\sqrt{n}}\right\|_2 \left\|\frac{\mathbf{\Sigma}_{\bar{r}} \mathbf{V}^\top \Delta\mathbf{w}}{\sqrt{n}}\right\|_2$$

$$\geq \left(\frac{1}{2} - \frac{\epsilon_1}{2}\right) \left\|\frac{\mathbf{\Sigma}_r \mathbf{V}^\top \Delta\mathbf{w}}{\sqrt{n}}\right\|_2^2 + \frac{\lambda n}{2\sigma_{r+1}^2}\left\|\frac{\mathbf{\Sigma}_{\bar{r}} \mathbf{V}^\top \Delta\mathbf{w}}{\sqrt{n}}\right\|_2^2$$

Consider the term $\left\langle \left(\mathbf{\Pi}\mathbf{\Pi}^\top - \mathbf{I}_n\right) \frac{\mathbf{X}\mathbf{w}^*}{\sqrt{n}}, -\frac{\mathbf{X}\Delta\mathbf{w}}{\sqrt{n}}\right\rangle$, we have

$$\left\langle \left(\mathbf{\Pi}\mathbf{\Pi}^\top - \mathbf{I}_n\right) \frac{\mathbf{X}\mathbf{w}^*}{\sqrt{n}}, \frac{\mathbf{X}}{\sqrt{n}}(\widehat{\mathbf{w}}_{\text{HS}} - \mathbf{w}^*)\right\rangle = \left\langle \left(\mathbf{\Pi}\mathbf{\Pi}^\top - \mathbf{I}_n\right) \frac{\mathbf{X}_r \mathbf{w}^*}{\sqrt{n}}, -\frac{\mathbf{X}_r \Delta\mathbf{w}}{\sqrt{n}}\right\rangle$$

$$+ \left\langle \left(\mathbf{\Pi}\mathbf{\Pi}^\top - \mathbf{I}_n\right) \frac{\mathbf{X}_{\bar{r}} \mathbf{w}^*}{\sqrt{n}}, -\frac{\mathbf{X}_r \Delta\mathbf{w}}{\sqrt{n}}\right\rangle$$

$$+ \left\langle \left(\mathbf{\Pi}\mathbf{\Pi}^\top - \mathbf{I}_n\right) \frac{\mathbf{X}_r \mathbf{w}^*}{\sqrt{n}}, -\frac{\mathbf{X}_{\bar{r}} \Delta\mathbf{w}}{\sqrt{n}}\right\rangle$$

$$+ \left\langle \left(\mathbf{\Pi}\mathbf{\Pi}^\top - \mathbf{I}_n\right) \frac{\mathbf{X}_{\bar{r}} \mathbf{w}^*}{\sqrt{n}}, -\frac{\mathbf{X}_{\bar{r}} \Delta\mathbf{w}}{\sqrt{n}}\right\rangle$$

.

Notice that the random matrix $\mathbf{\Pi}^\top \mathbf{U}_r$ and $\mathbf{\Pi}^\top \mathbf{U}_r$ can be treated as two Gaussian random matrices with entries smapled i.i.d from $\mathcal{N}(0, 1/m)$, applying Lemma A.1 and A.2 we can bound above terms separately:

$$\left\langle \left(\mathbf{\Pi}\mathbf{\Pi}^\top - \mathbf{I}_n\right) \frac{\mathbf{X}_r \mathbf{w}^*}{\sqrt{n}}, -\frac{\mathbf{X}_r \Delta\mathbf{w}}{\sqrt{n}}\right\rangle \leq \epsilon_1 \left\|\frac{\mathbf{\Sigma}_r \mathbf{V}^\top \mathbf{w}^*}{\sqrt{n}}\right\|_2 \left\|\frac{\mathbf{\Sigma}_r \mathbf{V}^\top \Delta\mathbf{w}}{\sqrt{n}}\right\|_2,$$

$$\left\langle \left(\mathbf{\Pi}\mathbf{\Pi}^\top - \mathbf{I}_n\right) \frac{\mathbf{X}_{\bar{r}} \mathbf{w}^*}{\sqrt{n}}, -\frac{\mathbf{X}_r \Delta\mathbf{w}}{\sqrt{n}}\right\rangle \leq \tau_1 \left\|\frac{\mathbf{\Sigma}_{\bar{r}} \mathbf{V}^\top \mathbf{w}^*}{\sqrt{n}}\right\|_2 \left\|\frac{\mathbf{\Sigma}_r \mathbf{V}^\top \Delta\mathbf{w}}{\sqrt{n}}\right\|_2,$$

$$\left\langle \left(\mathbf{\Pi}\mathbf{\Pi}^\top - \mathbf{I}_n\right) \frac{\mathbf{X}_r \mathbf{w}^*}{\sqrt{n}}, -\frac{\mathbf{X}_{\bar{r}} \Delta\mathbf{w}}{\sqrt{n}}\right\rangle \leq \tau_1 \left\|\frac{\mathbf{\Sigma}_r \mathbf{V}^\top \mathbf{w}^*}{\sqrt{n}}\right\|_2 \left\|\frac{\mathbf{\Sigma}_{\bar{r}} \mathbf{V}^\top \Delta\mathbf{w}}{\sqrt{n}}\right\|_2,$$

$$\left\langle \left(\mathbf{\Pi}\mathbf{\Pi}^\top - \mathbf{I}_n\right) \frac{\mathbf{X}_{\bar{r}} \mathbf{w}^*}{\sqrt{n}}, -\frac{\mathbf{X}_{\bar{r}} \Delta\mathbf{w}}{\sqrt{n}}\right\rangle \leq v_1 \left\|\frac{\mathbf{\Sigma}_{\bar{r}} \mathbf{V}^\top \mathbf{w}^*}{\sqrt{n}}\right\|_2 \left\|\frac{\mathbf{\Sigma}_{\bar{r}} \mathbf{V}^\top \Delta\mathbf{w}}{\sqrt{n}}\right\|_2.$$

By Cauchy-Schwarz inequality, we have

$$\left\langle \left(\mathbf{\Pi\Pi}^\top - \mathbf{I}_n\right) \frac{\mathbf{Xw}^*}{\sqrt{n}}, \frac{\mathbf{X}}{\sqrt{n}}(\widehat{\mathbf{w}}_{\mathrm{HS}} - \mathbf{w}^*) \right\rangle \leq \epsilon_1 \left\|\frac{\mathbf{\Sigma}_r \mathbf{V}^\top \mathbf{w}^*}{\sqrt{n}}\right\|_2 \left\|\frac{\mathbf{\Sigma}_r \mathbf{V}^\top \Delta\mathbf{w}}{\sqrt{n}}\right\|_2 + \tau_1 \left\|\frac{\mathbf{\Sigma}_{\bar{r}} \mathbf{V}^\top \mathbf{w}^*}{\sqrt{n}}\right\|_2 \left\|\frac{\mathbf{\Sigma}_r \mathbf{V}^\top \Delta\mathbf{w}}{\sqrt{n}}\right\|_2$$
$$+ \tau_1 \left\|\frac{\mathbf{\Sigma}_r \mathbf{V}^\top \mathbf{w}^*}{\sqrt{n}}\right\|_2 \left\|\frac{\mathbf{\Sigma}_{\bar{r}} \mathbf{V}^\top \Delta\mathbf{w}}{\sqrt{n}}\right\|_2 + \upsilon_1 \left\|\frac{\mathbf{\Sigma}_{\bar{r}} \mathbf{V}^\top \mathbf{w}^*}{\sqrt{n}}\right\|_2 \left\|\frac{\mathbf{\Sigma}_{\bar{r}} \mathbf{V}^\top \Delta\mathbf{w}}{\sqrt{n}}\right\|_2$$
$$\leq \frac{4\epsilon_1^2}{1-\epsilon_1}\left\|\frac{\mathbf{\Sigma}_r \mathbf{V}^\top \mathbf{w}^*}{\sqrt{n}}\right\|_2^2 + \frac{1-\epsilon_1}{8}\left\|\frac{\mathbf{\Sigma}_r \mathbf{V}^\top \Delta\mathbf{w}}{\sqrt{n}}\right\|_2^2$$
$$+ \frac{4\tau_1^2}{1-\epsilon_1}\left\|\frac{\mathbf{\Sigma}_{\bar{r}} \mathbf{V}^\top \mathbf{w}^*}{\sqrt{n}}\right\|_2^2 + \frac{1-\epsilon_1}{8}\left\|\frac{\mathbf{\Sigma}_r \mathbf{V}^\top \Delta\mathbf{w}}{\sqrt{n}}\right\|_2^2$$
$$+ \frac{4\tau_1^2 \sigma_{r+1}^2}{\lambda n}\left\|\frac{\mathbf{\Sigma}_r \mathbf{V}^\top \mathbf{w}^*}{\sqrt{n}}\right\|_2^2 + \frac{\lambda n}{8\sigma_{r+1}^2}\left\|\frac{\mathbf{\Sigma}_{\bar{r}} \mathbf{V}^\top \Delta\mathbf{w}}{\sqrt{n}}\right\|_2^2$$
$$+ \frac{4\upsilon_1^2 \sigma_{r+1}^2}{\lambda n}\left\|\frac{\mathbf{\Sigma}_{\bar{r}} \mathbf{V}^\top \mathbf{w}^*}{\sqrt{n}}\right\|_2^2 + \frac{\lambda n}{8\sigma_{r+1}^2}\left\|\frac{\mathbf{\Sigma}_{\bar{r}} \mathbf{V}^\top \Delta\mathbf{w}}{\sqrt{n}}\right\|_2^2$$

From the proof of 5.2 we know

$$\frac{1-\epsilon_1}{2}\left\|\frac{\mathbf{\Sigma}_r \mathbf{V}^\top \Delta\mathbf{w}}{\sqrt{n}}\right\|_2^2 + \frac{\lambda n}{2\sigma_{r+1}^2}\left\|\frac{\mathbf{\Sigma}_{\bar{r}} \mathbf{V}^\top \Delta\mathbf{w}}{\sqrt{n}}\right\|_2^2 \leq \Delta\mathbf{w}^\top \left(\frac{\mathbf{X}^\top \mathbf{\Pi\Pi}^\top \mathbf{X}}{n} + \lambda \mathbf{I}_p\right) \Delta\mathbf{w}$$
$$= \left\langle \left(\mathbf{\Pi\Pi}^\top - \mathbf{I}_n\right) \frac{\mathbf{Xw}^*}{\sqrt{n}}, \frac{\mathbf{X}}{\sqrt{n}}(\widehat{\mathbf{w}}_{\mathrm{HS}} - \mathbf{w}^*) \right\rangle$$

Combining above we have

$$\frac{1-\epsilon_1}{4}\left\|\frac{\mathbf{\Sigma}_r \mathbf{V}^\top \Delta\mathbf{w}}{\sqrt{n}}\right\|_2^2 + \frac{\lambda n}{4\sigma_{r+1}^2}\left\|\frac{\mathbf{\Sigma}_{\bar{r}} \mathbf{V}^\top \Delta\mathbf{w}}{\sqrt{n}}\right\|_2^2 \leq \left(\frac{4\epsilon_1^2}{1-\epsilon_1} + \frac{4\tau_1^2 \sigma_{r+1}^2}{\lambda n}\right)\left\|\frac{\mathbf{\Sigma}_r \mathbf{V}^\top \mathbf{w}^*}{\sqrt{n}}\right\|_2^2$$
$$+ \left(\frac{4\tau_1^2}{1-\epsilon_1} + \frac{4\upsilon_1^2 \sigma_{r+1}^2}{\lambda n}\right)\left\|\frac{\mathbf{\Sigma}_{\bar{r}} \mathbf{V}^\top \mathbf{w}^*}{\sqrt{n}}\right\|_2^2$$
$$\leq \left(\frac{4\epsilon_1^2}{1-\epsilon_1} + \frac{4\tau_1^2 \sigma_{r+1}^2}{\lambda n} + \frac{4\tau_1^2 \rho^2}{1-\epsilon} + \frac{4\rho^2 \upsilon_1^2 \sigma_{r+1}^2}{\lambda n}\right) \|\mathbf{w}^*\|_\mathbf{X}^2.$$

Thus

$$\|\Delta\mathbf{w}\|_\mathbf{X}^2 = \left\|\frac{\mathbf{\Sigma}_r \mathbf{V}^\top \Delta\mathbf{w}}{\sqrt{n}}\right\|_2^2 + \left\|\frac{\mathbf{\Sigma}_{\bar{r}} \mathbf{V}^\top \Delta\mathbf{w}}{\sqrt{n}}\right\|_2^2$$
$$\leq \left(\frac{4}{1-\epsilon_1} + \frac{4\sigma_{r+1}^2}{\lambda n}\right)\left(\frac{1-\epsilon_1}{4}\left\|\frac{\mathbf{\Sigma}_r \mathbf{V}^\top \Delta\mathbf{w}}{\sqrt{n}}\right\|_2^2 + \frac{\lambda n}{4\sigma_{r+1}^2}\left\|\frac{\mathbf{\Sigma}_{\bar{r}} \mathbf{V}^\top \Delta\mathbf{w}}{\sqrt{n}}\right\|_2^2\right)$$
$$\leq \left(\frac{4}{1-\epsilon_1} + \frac{4\sigma_{r+1}^2}{\lambda n}\right)\left(\frac{4\epsilon_1^2}{1-\epsilon_1} + \frac{4\tau_1^2 \sigma_{r+1}^2}{\lambda n} + \frac{4\tau_1^2 \rho^2}{1-\epsilon_1} + \frac{4\rho^2 \upsilon_1^2 \sigma_{r+1}^2}{\lambda n}\right)\|\mathbf{w}^*\|_\mathbf{X}^2,$$

which concludes the proof. $\square$

## A.3 Proof of Theorem 5.6

*Proof.* For notation simplicity let

$$\widetilde{\mathbf{H}} = \frac{\mathbf{X}^\top \mathbf{\Pi}\mathbf{\Pi}^\top \mathbf{X}}{n} + \lambda \mathbf{I}_p \quad \text{and} \quad \mathbf{H} = \frac{\mathbf{X}^\top \mathbf{X}}{n} + \lambda \mathbf{I}_p,$$

based on the property of similarity matrices, we have

$$\kappa(\widetilde{\mathbf{H}}^{-1}\mathbf{H}) = \kappa(\widetilde{\mathbf{H}}^{-1/2}\mathbf{H}\widetilde{\mathbf{H}}^{-1/2}) = \frac{\max_{\mathbf{w}} \mathbf{w}^\top \widetilde{\mathbf{H}}^{-1/2}\mathbf{H}\widetilde{\mathbf{H}}^{-1/2}\mathbf{w}}{\min_{\mathbf{w}} \mathbf{w}^\top \widetilde{\mathbf{H}}^{-1/2}\mathbf{H}\widetilde{\mathbf{H}}^{-1/2}\mathbf{w}}.$$

Consider the quantity $|\mathbf{w}^\top \widetilde{\mathbf{H}}^{-1/2}(\mathbf{H}-\widetilde{\mathbf{H}})\widetilde{\mathbf{H}}^{-1/2}\mathbf{w}|$, since

$$\begin{aligned}
|\mathbf{w}^\top \widetilde{\mathbf{H}}^{-1/2}(\mathbf{H}-\widetilde{\mathbf{H}})\widetilde{\mathbf{H}}^{-1/2}\mathbf{w}| &= \left\langle \left(\mathbf{H}-\widetilde{\mathbf{H}}\right)\widetilde{\mathbf{H}}^{-1/2}\mathbf{w}, \widetilde{\mathbf{H}}^{-1/2}\mathbf{w} \right\rangle \\
&= \left\langle \left(\mathbf{\Pi}\mathbf{\Pi}^\top - \mathbf{I}_n\right)\frac{\mathbf{X}}{\sqrt{n}}\widetilde{\mathbf{H}}^{-1/2}\mathbf{w}, \frac{\mathbf{X}}{\sqrt{n}}\widetilde{\mathbf{H}}^{-1/2}\mathbf{w} \right\rangle \\
&\leqslant \rho_2\left(\mathbf{X}\mathbb{R}^p, \mathbf{\Pi}, \frac{\mathbf{X}}{\sqrt{n}}\widetilde{\mathbf{H}}^{-1/2}\mathbf{w}\right) \left\|\widetilde{\mathbf{H}}^{-1/2}\mathbf{w}\right\|_{\mathbf{X}}^2 \\
&\leqslant C_0 \sqrt{\frac{\mathbb{W}^2(\mathbf{X}\mathbb{R}^p)}{m}\log\left(\frac{1}{\delta}\right)} \left\|\widetilde{\mathbf{H}}^{-1/2}\mathbf{w}\right\|_{\mathbf{X}}^2
\end{aligned}$$

Since for any vector $\mathbf{u} \in \mathbb{R}^p$, we have

$$\begin{aligned}
\left\|\widetilde{\mathbf{H}}^{1/2}\mathbf{u}\right\|_2^2 &= \mathbf{u}^\top \left(\frac{\mathbf{X}^\top \mathbf{\Pi}\mathbf{\Pi}^\top \mathbf{X}}{n} + \lambda \mathbf{I}_p\right)\mathbf{u} \\
&= \mathbf{u}^\top \left(\frac{\mathbf{X}^\top \mathbf{\Pi}\mathbf{\Pi}^\top \mathbf{X}}{n}\right)\mathbf{u} + \lambda \|\mathbf{u}\|_2^2 \\
&\geqslant \rho_1(\mathbf{X}\mathbb{R}^p, \mathbf{\Pi}) \|\mathbf{u}\|_{\mathbf{X}}^2 \\
&\geqslant \left(1 - C_0\sqrt{\frac{\mathbb{W}^2(\mathbf{X}\mathbb{R}^p)}{m}\log\left(\frac{1}{\delta}\right)}\right) \|\mathbf{u}\|_{\mathbf{X}}^2.
\end{aligned}$$

Let $\mathbf{u} = \widetilde{\mathbf{H}}^{-1/2}\mathbf{w}$ we have

$$\left\|\widetilde{\mathbf{H}}^{-1/2}\mathbf{w}\right\|_{\mathbf{X}}^2 \leqslant \frac{1}{1 - C_0\sqrt{\frac{\mathbb{W}^2(\mathbf{X}\mathbb{R}^p)}{m}\log\left(\frac{1}{\delta}\right)}} \|\mathbf{w}\|_2^2$$

Combining above we get

$$|\mathbf{w}^\top \widetilde{\mathbf{H}}^{-1/2}(\mathbf{H}-\widetilde{\mathbf{H}})\widetilde{\mathbf{H}}^{-1/2}\mathbf{w}| \leqslant \frac{C_0\sqrt{\frac{\mathbb{W}^2(\mathbf{X}\mathbb{R}^p)}{m}\log\left(\frac{1}{\delta}\right)}}{1 - C_0\sqrt{\frac{\mathbb{W}^2(\mathbf{X}\mathbb{R}^p)}{m}\log\left(\frac{1}{\delta}\right)}} \|\mathbf{w}\|_2^2,$$

which implies

$$\begin{aligned}
\max_{\mathbf{w}} \mathbf{w}^\top \widetilde{\mathbf{H}}^{-1/2}\mathbf{H}\widetilde{\mathbf{H}}^{-1/2}\mathbf{w} &\leqslant \|\mathbf{w}\|_2^2 + \frac{C_0\sqrt{\frac{\mathbb{W}^2(\mathbf{X}\mathbb{R}^p)}{m}\log\left(\frac{1}{\delta}\right)}}{1 - C_0\sqrt{\frac{\mathbb{W}^2(\mathbf{X}\mathbb{R}^p)}{m}\log\left(\frac{1}{\delta}\right)}} \|\mathbf{w}\|_2^2 \\
&= \frac{1}{1 - C_0\sqrt{\frac{\mathbb{W}^2(\mathbf{X}\mathbb{R}^p)}{m}\log\left(\frac{1}{\delta}\right)}} \|\mathbf{w}\|_2^2,
\end{aligned}$$

and
$$\min_{\mathbf{w}} \mathbf{w}^\top \widetilde{\mathbf{H}}^{-1/2} \mathbf{H} \widetilde{\mathbf{H}}^{-1/2} \mathbf{w} \geqslant \|\mathbf{w}\|_2^2 - \frac{C_0 \sqrt{\frac{\mathbb{W}^2(\mathbf{X}\mathbb{R}^p)}{m} \log\left(\frac{1}{\delta}\right)}}{1 - C_0 \sqrt{\frac{\mathbb{W}^2(\mathbf{X}\mathbb{R}^p)}{m} \log\left(\frac{1}{\delta}\right)}} \|\mathbf{w}\|_2^2$$
$$= \frac{1 - 2C_0 \sqrt{\frac{\mathbb{W}^2(\mathbf{X}\mathbb{R}^p)}{m} \log\left(\frac{1}{\delta}\right)}}{1 - C_0 \sqrt{\frac{\mathbb{W}^2(\mathbf{X}\mathbb{R}^p)}{m} \log\left(\frac{1}{\delta}\right)}} \|\mathbf{w}\|_2^2,$$

Thus we know
$$\kappa(\widetilde{\mathbf{H}}^{-1}\mathbf{H}) \leqslant \frac{1}{1 - 2C_0 \sqrt{\frac{\mathbb{W}^2(\mathbf{X}\mathbb{R}^p)}{m} \log\left(\frac{1}{\delta}\right)}},$$
and proof for $\kappa_{\mathrm{DRP}}(\mathbf{X}, \mathbf{R}, \lambda)$ is analogous. $\square$

### A.4 Proof of Lemma 5.8

*Proof.* Note that (2.6) is sketching the following problem:
$$\arg\min_{\mathbf{u}} \mathbf{u}^\top \left(\frac{\mathbf{X}^\top \mathbf{X}}{2n} + \frac{\lambda}{2}\mathbf{I}_p\right) \mathbf{u} - \left\langle \frac{\mathbf{X}^\top(\mathbf{y} - \mathbf{X}\widehat{\mathbf{w}}_{\mathrm{HS}}^{(t)})}{n} - \lambda \widehat{\mathbf{w}}_{\mathrm{HS}}^{(t)}, \mathbf{u} \right\rangle,$$
where $\mathbf{w}^* - \widehat{\mathbf{w}}_{\mathrm{HS}}^{(t)}$ is the optimal solution, thus applying Theorem 5.2, we have
$$\left\|\widehat{\mathbf{u}}^{(t)} - (\mathbf{w}^* - \widehat{\mathbf{w}}_{\mathrm{HS}}^{(t)})\right\|_{\mathbf{X}} \leqslant \frac{C_0 \sqrt{\frac{\mathbb{W}^2(\mathbf{X}\mathbb{R}^p)}{m} \log\left(\frac{1}{\delta}\right)}}{1 - C_0 \sqrt{\frac{\mathbb{W}^2(\mathbf{X}\mathbb{R}^p)}{m} \log\left(\frac{1}{\delta}\right)}} \left\|\widehat{\mathbf{w}}_{\mathrm{HS}}^{(t)} - \mathbf{w}^*\right\|_{\mathbf{X}},$$
using the definition that $\widehat{\mathbf{w}}_{\mathrm{HS}}^{(t+1)} = \widehat{\mathbf{w}}_{\mathrm{HS}}^{(t)} + \widehat{\mathbf{u}}^{(t)}$, we obtain the desired result. $\square$

### A.5 Proof of Theorem 5.9

*Proof.* By triangle inequality we have the following decomposition:
$$\left\|\widetilde{\mathbf{w}}_{\mathrm{HS}}^{(t+1)} - \mathbf{w}^*\right\|_{\mathbf{X}} \leqslant \left\|\widehat{\mathbf{w}}_{\mathrm{HS}}^{(t+1)} - \mathbf{w}^*\right\|_{\mathbf{X}} + \left\|\widetilde{\mathbf{w}}_{\mathrm{HS}}^{(t+1)} - \widehat{\mathbf{w}}_{\mathrm{HS}}^{(t+1)}\right\|_{\mathbf{X}}$$
$$\leqslant \frac{C_0 \sqrt{\frac{\mathbb{W}^2(\mathbf{X}\mathbb{R}^p)}{m} \log\left(\frac{1}{\delta}\right)}}{1 - C_0 \sqrt{\frac{\mathbb{W}^2(\mathbf{X}\mathbb{R}^p)}{m} \log\left(\frac{1}{\delta}\right)}} \left\|\widehat{\mathbf{w}}_{\mathrm{HS}}^{(t)} - \mathbf{w}^*\right\|_{\mathbf{X}} + \left\|\widetilde{\mathbf{w}}_{\mathrm{HS}}^{(t+1)} - \widehat{\mathbf{w}}_{\mathrm{HS}}^{(t+1)}\right\|_{\mathbf{X}}$$
$$\leqslant \left(\frac{C_0 \sqrt{\frac{\mathbb{W}^2(\mathbf{X}\mathbb{R}^p)}{m} \log\left(\frac{1}{\delta}\right)}}{1 - C_0 \sqrt{\frac{\mathbb{W}^2(\mathbf{X}\mathbb{R}^p)}{m} \log\left(\frac{1}{\delta}\right)}}\right)^t \|\mathbf{w}^*\|_{\mathbf{X}} + \left\|\widetilde{\mathbf{w}}_{\mathrm{HS}}^{(t+1)} - \widehat{\mathbf{w}}_{\mathrm{HS}}^{(t+1)}\right\|_{\mathbf{X}}$$

For the term $\left\|\widetilde{\mathbf{w}}_{\mathrm{HS}}^{(t+1)} - \widehat{\mathbf{w}}_{\mathrm{HS}}^{(t+1)}\right\|_{\mathbf{X}}$, we can further bridge $\widetilde{\mathbf{w}}_{\mathrm{HS}}^{(t+1)}$ and $\widehat{\mathbf{w}}_{\mathrm{HS}}^{(t+1)}$ by $\bar{\mathbf{w}}_{\mathrm{HS}}^{(t+1)}$, which is the result of one exact step of IHS initialized at $\widetilde{\mathbf{w}}_{\mathrm{HS}}^{(t)}$. Thus we have the following decomposition
$$\left\|\widetilde{\mathbf{w}}_{\mathrm{HS}}^{(t+1)} - \widehat{\mathbf{w}}_{\mathrm{HS}}^{(t+1)}\right\|_{\mathbf{X}} \leqslant \left\|\widetilde{\mathbf{w}}_{\mathrm{HS}}^{(t+1)} - \bar{\mathbf{w}}_{\mathrm{HS}}^{(t+1)}\right\|_{\mathbf{X}} + \left\|\bar{\mathbf{w}}_{\mathrm{HS}}^{(t+1)} - \widehat{\mathbf{w}}_{\mathrm{HS}}^{(t+1)}\right\|_{\mathbf{X}}$$

Applying the Theorem 5.2 for DRP we have the following bound for $\left\|\widetilde{\mathbf{w}}_{\text{HS}}^{(t+1)} - \bar{\mathbf{w}}_{\text{HS}}^{(t+1)}\right\|_{\mathbf{X}}$:

$$\left\|\widetilde{\mathbf{w}}_{\text{HS}}^{(t+1)} - \bar{\mathbf{w}}_{\text{HS}}^{(t+1)}\right\|_{\mathbf{X}} \leq \lambda_{\max}\left(\frac{\mathbf{X}^\top\mathbf{X}}{n}\right) \left\|\widetilde{\mathbf{w}}_{\text{HS}}^{(t+1)} - \bar{\mathbf{w}}_{\text{HS}}^{(t+1)}\right\|_2$$

$$\leq \lambda_{\max}\left(\frac{\mathbf{X}^\top\mathbf{X}}{n}\right) \left(\frac{C_0\sqrt{\frac{\mathbb{W}^2(\mathbf{X}^\top\mathbb{R}^n)}{d}}\log\left(\frac{1}{\delta}\right)}{1 - C_0\sqrt{\frac{\mathbb{W}^2(\mathbf{X}^\top\mathbb{R}^n)}{d}}\log\left(\frac{1}{\delta}\right)}\right)^k \left\|\bar{\mathbf{w}}_{\text{HS}}^{(t+1)}\right\|_2$$

$$\leq \lambda_{\max}\left(\frac{\mathbf{X}^\top\mathbf{X}}{n}\right) \left(\frac{C_0\sqrt{\frac{\mathbb{W}^2(\mathbf{X}^\top\mathbb{R}^n)}{d}}\log\left(\frac{1}{\delta}\right)}{1 - C_0\sqrt{\frac{\mathbb{W}^2(\mathbf{X}^\top\mathbb{R}^n)}{d}}\log\left(\frac{1}{\delta}\right)}\right)^k (\left\|\bar{\mathbf{w}}_{\text{HS}}^{(t+1)} - \mathbf{w}^*\right\|_2 + \|\mathbf{w}^*\|_2)$$

$$\leq 2\lambda_{\max}\left(\frac{\mathbf{X}^\top\mathbf{X}}{n}\right) \left(\frac{C_0\sqrt{\frac{\mathbb{W}^2(\mathbf{X}^\top\mathbb{R}^n)}{d}}\log\left(\frac{1}{\delta}\right)}{1 - C_0\sqrt{\frac{\mathbb{W}^2(\mathbf{X}^\top\mathbb{R}^n)}{d}}\log\left(\frac{1}{\delta}\right)}\right)^k \|\mathbf{w}^*\|_2$$

Also, we can relate the error $\left\|\bar{\mathbf{w}}_{\text{HS}}^{(t+1)} - \widehat{\mathbf{w}}_{\text{HS}}^{(t+1)}\right\|_{\mathbf{X}}$ to the error term at $t$-th outer loop iteration: $\left\|\widetilde{\mathbf{w}}_{\text{HS}}^{(t)} - \widehat{\mathbf{w}}_{\text{HS}}^{(t)}\right\|_{\mathbf{X}}$:

$$\left\|\bar{\mathbf{w}}_{\text{HS}}^{(t+1)} - \widehat{\mathbf{w}}_{\text{HS}}^{(t+1)}\right\|_{\mathbf{X}} = \left\|\widetilde{\mathbf{w}}_{\text{HS}}^{(t)} - \widetilde{\mathbf{H}}^{-1}\nabla P(\widetilde{\mathbf{w}}_{\text{HS}}^{(t)}) - \widehat{\mathbf{w}}_{\text{HS}}^{(t)} - \widetilde{\mathbf{H}}^{-1}\nabla P(\widehat{\mathbf{w}}_{\text{HS}}^{(t)})\right\|_{\mathbf{X}}$$

$$= \left\|\widetilde{\mathbf{H}}^{-1}(\widetilde{\mathbf{H}} - \mathbf{H})(\widetilde{\mathbf{w}}_{\text{HS}}^{(t)} - \widehat{\mathbf{w}}_{\text{HS}}^{(t)})\right\|_{\mathbf{X}}$$

$$\leq \left\|\widetilde{\mathbf{H}}^{-1}\right\|_2 \left\|\widetilde{\mathbf{H}} - \mathbf{H}\right\|_2 \left\|\widetilde{\mathbf{w}}_{\text{HS}}^{(t)} - \widehat{\mathbf{w}}_{\text{HS}}^{(t)}\right\|_{\mathbf{X}}$$

$$\leq \frac{4\lambda_{\max}\left(\frac{\mathbf{X}^\top\mathbf{X}}{n}\right)}{\lambda} \left\|\widetilde{\mathbf{w}}_{\text{HS}}^{(t)} - \widehat{\mathbf{w}}_{\text{HS}}^{(t)}\right\|_{\mathbf{X}}$$

$$\leq \frac{8\lambda_{\max}^2\left(\frac{\mathbf{X}^\top\mathbf{X}}{n}\right)}{\lambda} \left(\frac{C_0\sqrt{\frac{\mathbb{W}^2(\mathbf{X}^\top\mathbb{R}^n)}{d}}\log\left(\frac{1}{\delta}\right)}{1 - C_0\sqrt{\frac{\mathbb{W}^2(\mathbf{X}^\top\mathbb{R}^n)}{d}}\log\left(\frac{1}{\delta}\right)}\right)^k \|\mathbf{w}^*\|_2$$

Combining above inequalities we obtained the following iterative error bound for $\widetilde{\mathbf{w}}_{\text{HS}}^{(t+1)}$:

$$\left\|\widetilde{\mathbf{w}}_{\text{HS}}^{(t+1)} - \mathbf{w}^*\right\|_{\mathbf{X}} \leq \left(\frac{C_0\sqrt{\frac{\mathbb{W}^2(\mathbf{X}\mathbb{R}^p)}{m}}\log\left(\frac{1}{\delta}\right)}{1 - C_0\sqrt{\frac{\mathbb{W}^2(\mathbf{X}\mathbb{R}^p)}{m}}\log\left(\frac{1}{\delta}\right)}\right)^t \|\mathbf{w}^*\|_{\mathbf{X}}$$

$$+ \frac{10\lambda_{\max}^2\left(\frac{\mathbf{X}^\top\mathbf{X}}{n}\right)}{\lambda} \left(\frac{C_0\sqrt{\frac{\mathbb{W}^2(\mathbf{X}^\top\mathbb{R}^n)}{d}}\log\left(\frac{1}{\delta}\right)}{1 - C_0\sqrt{\frac{\mathbb{W}^2(\mathbf{X}^\top\mathbb{R}^n)}{d}}\log\left(\frac{1}{\delta}\right)}\right)^k \|\mathbf{w}^*\|_2$$

$\square$